\documentclass{article}

\usepackage[preprint,nonatbib]{neurips_2021}

\usepackage[utf8]{inputenc} 
\usepackage[T1]{fontenc}    
\usepackage{hyperref}       
\usepackage{url}            
\usepackage{booktabs}       
\usepackage{amsfonts}       
\usepackage{nicefrac}       
\usepackage{microtype}      
\usepackage{xcolor}         

\usepackage{graphicx}
\usepackage{bm}
\usepackage{amsthm}
\usepackage{amsmath}
\usepackage{mathrsfs}
\usepackage{hyperref}
\usepackage{mathtools}
\hypersetup{
    colorlinks=true,
    linkcolor=blue,
    filecolor=magenta,      
    urlcolor=cyan,
}
\usepackage{subcaption}

\title{Kinematically consistent recurrent neural networks for learning inverse problems in wave propagation}

\author{%
  Wrik Mallik\\
  Department of Mechanical Engineering\\
  The University of British Columbia\\
  Vancouver, BC V6T 1Z4, Canada\\
  \texttt{wrik.mallik@ubc.ca}\\
   \And
  Rajeev K. Jaiman \\
  Department of Mechanical Engineering\\
  The University of British Columbia\\
  Vancouver, BC V6T 1Z4, Canada\\
  \texttt{rjaiman@mech.ubc.ca}\\
   \And
  Jasmin Jelovica \\
  Department of Mechanical Engineering\\
  The University of British Columbia\\
  Vancouver, BC V6T 1Z4, Canada\\
  \texttt{jjelovica@mech.ubc.ca}\\
}

\begin{document}

\maketitle

\begin{abstract}
   Although machine learning (ML) is increasingly employed recently for mechanistic problems, the black-box nature of conventional ML architectures lacks the physical knowledge to infer unforeseen input conditions. This implies both severe overfitting during a dearth of training data and inadequate physical interpretability, which motivates us to propose a new kinematically consistent, physics-based ML model. In particular, we attempt to perform physically interpretable learning of inverse problems in wave propagation without suffering overfitting restrictions.
  Towards this goal, we employ long short-term memory (LSTM) networks endowed with a physical, hyperparameter-driven regularizer, performing penalty-based enforcement of the characteristic geometries. Since these characteristics are the kinematical invariances of wave propagation phenomena, maintaining their structure provides kinematical consistency to the network.  
  Even with modest training data, the kinematically consistent network can reduce the $L_1$ and $L_\infty$ error norms of the plain LSTM predictions by about 45\% and 55\%, respectively. It can also increase the horizon of the plain LSTM's forecasting by almost two times. To achieve this, an optimal range of the physical hyperparameter, analogous to an artificial bulk modulus, has been established through numerical experiments. The efficacy of the proposed method in alleviating overfitting, and the physical interpretability of the learning mechanism, are also discussed. Such an application of kinematically consistent LSTM networks for wave propagation learning is presented here for the first time.
\end{abstract}

\section{Introduction}
In recent years, machine learning (ML) models have enjoyed tremendous success in a vast range of commercial applications from image processing, natural language processing, speech recognition to drug delivery and stock market prediction. This has led to their rapidly increasing usage in domains traditionally dominated by mechanistic models \cite{hsieh2009machine, bergen2019machine, reichstein2019deep}. The main motivation of using ML and deep learning (DL) models for mechanistic problems is their ability to induce certain biases to the model based on their training data. Such inductive bias allows ML/DL models to generalize on test data that follows the same distribution of the training data, even though these ML/DL models are over-parameterized in most cases and could easily lead to overfitting. It has been shown via recent numerical experiments \cite{mansour2019deep} that even if ML/DL models are trained with noisy data they tend to prioritize the learning of some low dimensional manifold consisting of certain simple features and only later tries to fit the complex behavior of the noisy signal. Although there is no general rule pertaining to the selection of the simple features or even classifying which features are simple features, it usually relies on either geometric priors provided by the ML/DL framework \cite{bronstein2021geometric}, or other priors like the activation functions employed in the ML/DL model. Such induction of simple laws governing the data aligns perfectly with data of mechanistic origin, which are governed by a few fundamental physical principles (e.g., Newton's laws of motion). This underscores the motivation for approximating mechanistic models with data-driven ML/DL models. 

Most problems of engineering importance are of complex, nonlinear and multiscale nature. In these problems, solutions to the governing partial differential equations (PDEs) are often analytically unavailable. On the other hand, numerical solution approaches like finite-difference methods (FDM), finite-element methods (FEM), etc., are restricted by the curse of dimensionality. In such situations, we often look towards data-driven dimensional reduction approaches to improve computational efficiency. Although linear projection-based methods like proper orthogonal decomposition (POD) are popularly used model reduction techniques, they can prove inefficient for physical problems where the worst-case error from the best-approximated linear subspace decays slowly with an increase in subspace dimension (i.e., large Kolmogorov N-width problem) \cite{taddei2020registration, mojgani2020physics, greif2019decay} because of the linear mapping between the large dimensional set of physical states and the much reduced dimensional set of POD latent states.

Dimensionality reduction for nonlinear physical phenomena, e.g. fluid-structure interaction around bluff bodies, can also be performed by various hyper-reduction methods like Discrete Empirical Interpolation Method (DEIM) \cite{miyanawala2019decomposition}, which is supposed to provide improved efficiency over linear projection methods. However, DEIM is still based on the POD reduced basis functions and complications can occur during the reconstruction of nonlinear flow features and forces. Besides, the basis functions of the POD-DEIM model must still satisfy the orthogonality restriction. Neural networks, the workhorse of ML/DL, on the other hand, are heuristic models which provide a nonlinear mapping between the high-dimensional data and lower-dimensional latent states via a flexible combination of several nonlinear functions. They are flexible yet optimal in nature, as linear activation functions will lead to similar results as POD. The nonlinear map provided by the neural networks obviates the large Kolmogorov N-width problem, and the flexibility of even the simplest of ML/DL architectures allows them to approximate many complex functions. This has been demonstrated by various universal approximation theorems (UATs) \cite{cybenko1989approximation, hornik1991approximation, leshno1993multilayer, pinkus1999approximation}. This has led to the employment of encoder-propagator-decoder networks for accurately learning various nonlinear and complex physical problems in a much lower dimension \cite{gonzalez2018deep, sorteberg2019approximating, cheng2020data, lee2020model, parish2020time, bukka2021assessment} and providing efficient reduced-order models (ROMs).

Although UATs demonstrate the approximation capabilities of ML/DL models, even the most sophisticated ML/DL architectures can only perform as a black-box model for physical problems and lack any physical information required to forecast physically consistent outputs for unforeseen input conditions. This has led to the development of physics-guided neural networks where domain-specific physical information is integrated with the ML architecture \cite{willard2020integrating}. In particular, these physics-guided networks can be classified into two main categories. The first category integrates physical information by applying a regularizer to the standard loss function. Such regularizers are based on physical conservation principles or the governing equations of the problems themselves and can be used for both improving predictions \cite{jia2019physics} or approximating PDEs \cite{raissi2019physics, yang2020physics}. The other category consists of modifying the ML/DL architecture to incorporate physical information by adding intermediate physical variables to conventional networks \cite{daw2020physics, sun2020theory}, by encoding invariances and symmetries within the architecture \cite{ling2016reynolds, wang2020incorporating, udrescu2020ai, ruthotto2019deep, greydanus2019hamiltonian, toth2019hamiltonian, wang2020towards}, and by providing other domain-specific physical knowledge, which does not correspond to known invariances or symmetries but provides meaningful structure to the optimization process \cite{wehmeyer2018time, otto2019linearly, chang2019antisymmetricrnn, ba2019blending}.

Here we are interested in learning inverse problems in wave propagation, which on a large scale is analogous to learning underwater radiated noise (URN) propagation originating from various human activities. Such URN propagation acts as a stressor for underwater marine animals \cite{duarte2021soundscape} and is a significant concern for both environmentalists and the shipbuilding industry. Machine learning of URN propagation can lead to both overfitting and underfitting of even sophisticated networks due to a dearth of high-quality training data. This motivates the application of physics-based, scalable learning techniques. Towards this goal, we propose an approach for developing physics-based long short-term memory (LSTM) networks for learning the inverse problem in wave propagation, as explained below.

\begin{figure}[ht]
    \centering
    \begin{subfigure}{.49\textwidth}
        \centering
        \includegraphics[width=\linewidth]{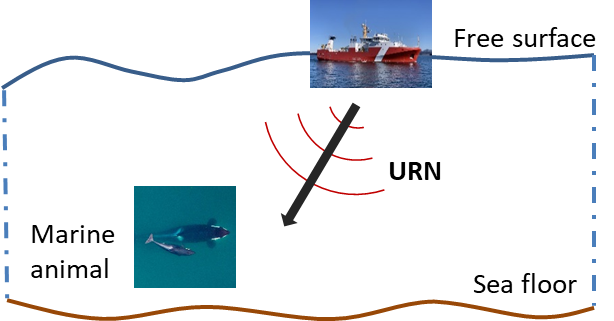}
        \caption{URN from marine vessels}
    \end{subfigure}
    \hfill
    \begin{subfigure}{.49\textwidth}
        \centering
        \includegraphics[width=\linewidth]{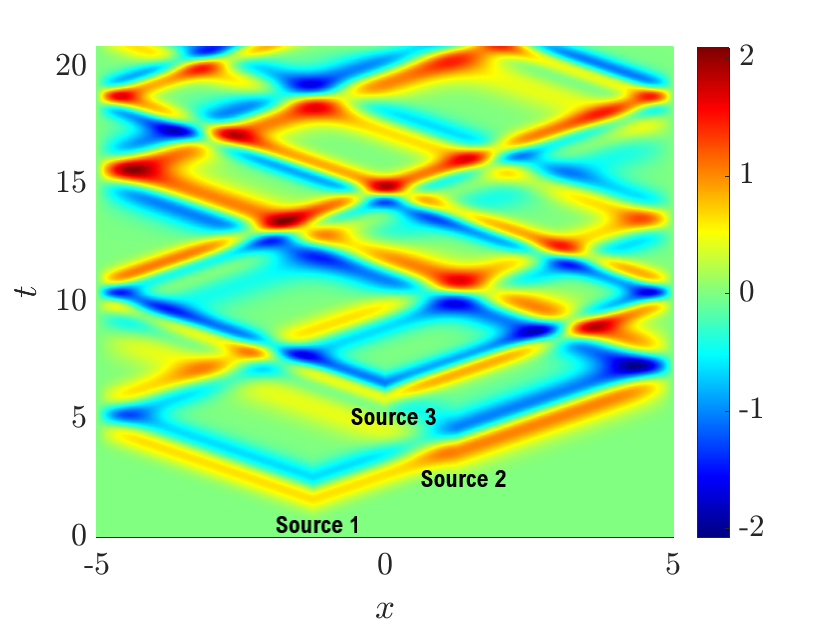}
        \caption{1-D multi-source wave characteristics}
    \end{subfigure}
    \caption{Simplified characteristic map of URN propagation from marine vessels}
    \label{fig:characteristics_anthropocene_ocean}
\end{figure}

We propose a kinematically consistent LSTM network where problem kinematics are induced to the network via wave characteristics. These characteristics form the propagation pathways for any general transport problem, including wave propagation. Wave characteristic patterns of a multi-source, one-dimensional realization of human-generated URN is shown in Fig. \ref{fig:characteristics_anthropocene_ocean}. Predictions of correct characteristic patterns can be achieved by minimizing errors in the predicted wave speed, thereby keeping the kinematical properties consistent. Thus, a physical regularizer minimizing the mean square error (MSE) between the target and predicted wave speed is added to the standard training loss function. The regularizer is scaled by a hyperparameter that resembles an artificial bulk modulus, thus providing physical context to the regularization process. The wave speed computation is completely data-driven without requiring any use of PDEs, unlike the approach adopted in some other physics-informed neural network architectures \cite{raissi2019physics}. To the best of the authors' knowledge, such application of kinematical consistency for inverse learning of mechanistic problems is novel in the physics-guided ML community, especially for wave propagation phenomena. The efficiency of the physics-based network in alleviating the overfitting of plain LSTM networks, interpretability of the physical regularization, and comparison of the forecasting capabilities of the physics-based and the plain LSTMs are discussed later in the article.

\section{Methodology}
\subsection{Underwater wave propagation as an inverse problem}
Underwater noise radiations can be considered as a wave propagation problem with the assumptions that the fluid is isotropic and homogeneous, viscous stresses are negligible, the process is adiabatic, and the spatial variations of the ambient pressure, density and temperature are relatively very small. Under these assumptions, physical quantities like density and pressure can be expressed as a sum of their steady values and unsteady fluctuations of much smaller amplitude. Using the aforementioned assumptions, the conservation of mass and momentum, and the equation of state of the fluid, the propagation of pressure fluctuations in a one-dimensional domain can be represented as a second-order hyperbolic PDE,
\begin{equation}
    \label{eq:acoustic_PDE_pressure}
    \frac{\partial^2 \acute{p}}{{\partial t}^2} - {c_o}^2 \nabla^2 \acute{p} = q(x,t;x_0,t_0), \quad x \in \left[-\frac{L}{2},\frac{L}{2}\right],
\end{equation}
subject to the initial and boundary conditions. Here, $q$ is a source term subject to the source location $x_0$, and initiation point $t_0$, $c_0$ is the constant speed of sound in the medium, and $L$ is the total domain length. Equation \ref{eq:acoustic_PDE_pressure} can also be expressed as,
\begin{equation}
    \label{eq:prob_interest}
    \bm{\mathcal{L}} \acute{p} = q(x,t;x_0,t_0), \quad x \in \left[-\frac{L}{2},\frac{L}{2}\right],
\end{equation}
where $\bm{\mathcal{L}} = \frac{\partial^2}{\partial t^2} - {c_o}^2 \nabla^2$, is the differential operator. Eq. (\ref{eq:prob_interest}) is usually solved as a forward problems from the causality (sources, initial and boundary conditions, etc.) to the effects (pressure fluctuations). For homogeneous, range-dependent ocean properties and uniform boundaries we can obtain closed-form integral solutions to Eq. \ref{eq:prob_interest} as,
\begin{multline}
    \label{eq:closed-Form_sol}
    \acute{p}(x,t) = \int_{t_0}^t \int_{x_0}^x q(y,\tau) \mathscr{G}(x,t;y,\tau) \, dy \, d\tau -  c_0^2 {\Bigg[ \int_{t_0}^t \left[\acute{p}(y,\tau) \frac{\partial \mathscr{G}}{\partial y} - \mathscr{G}\frac{\partial \acute{p}(y,\tau)}{\partial y}\right] \, d\tau \Bigg]}_{y=x_0} \\ 
    - {\Bigg[\int_{x_0}^x \left[\acute{p}(y,\tau) \frac{\partial \mathscr{G}}{\partial \tau} - \mathscr{G}\frac{\partial \acute{p}(y,\tau)}{\partial \tau}\right] \, dy \Bigg]}_{\tau=t_0},
\end{multline}
where the second term on the right-hand side represents the boundary conditions and the third term represents the initial conditions. $\mathscr{G}$ is the Green's function and is obtained from the homogeneous analog of Eq. (\ref{eq:prob_interest}),
\begin{equation}
    \label{eq:Greens_func}
    \bm{\mathcal{L}} \mathscr{G}(x,t;y,\tau) = \delta(x - y) \delta(t - \tau).
\end{equation}
Here, $\delta(x - y) \delta(t - \tau)$ is a pulse released at location $y$ and time $\tau$, and observed at location $x$ and time $t$. For complex cases with localized fluid properties or non-uniform boundaries, finding Green's functions is intractable. In such situations we can resort to numerical techniques like the FDM, FEM, spectral methods, etc., where we approximate $\bm{L}$ to numerically obtain our solutions. 

There are often situations for URN propagation where we encounter uncertainties or lack of physical knowledge of the actual phenomena. Under such circumstances, solving the forward problem may be restricted. Instead, we pose our problem as an inverse problem from the effects to the causality. The inverse map is learned by data-driven ML architectures, which formulate the inverse learning of the temporal evolution of waves as a sequence-to-sequence ML problem. 

Although commercially successful sequence-to-sequence ML architectures are available, these can only serve as a black-box data-driven model. We plan to induct domain-specific physical information to them for enhancing their capability and obtain more physically consistent predictions. The fundamental physical knowledge that is inducted to the black-box ML architectures can be inferred from the general solution of 1-D plane waves, as indicated by the Green's functions for such cases:
\begin{equation}
    \label{eq:Greens_function_1Dplanewaves}
    \mathscr{G}(x,t;y,\tau) = \mathscr{F}_1 \left(t-t_0 + \frac{x-x_0}{c_0}\right) + \mathscr{F}_2 \left(t-t_0 - \frac{x-x_0}{c_0}\right).
\end{equation}
Here, the exact form of the functions $\mathscr{F}_1$ and $\mathscr{F}_2$ $\big($combinations of \textit{Heaviside} functions of arguments $t-t_0 \pm \frac{x-x_0}{c_0}\big)$ depends on the boundary conditions and the phase of the right and left traveling reflected waves. However, the arguments of $\mathscr{F}_1$ and $\mathscr{F}_2$ are the characteristics of any linear, one-dimensional wave propagation phenomena and they can be used to describe the kinematical properties of the wave propagation. Such physical relation between the wave characteristics and the wave kinematics will be employed for endowing kinematical consistency to an LSTM network, a state-of-the-art sequence-to-sequence learning model.

\subsection{Kinematically consistent LSTM network}
\subsubsection{Recursive LSTM architecture}
LSTMs are gated recurrent neural networks (RNNs) routinely used for accurately learning sequences with a long-term data dependency. The gating mechanism of LSTMs provides them invariance to time warping. Thus they suffer negligible adverse effects due to vanishing gradients unlike the non-gated RNNs. The LSTM architecture used here comprises a standard LSTM cell with the input gate, the output gate, and the forget gate. The cell input, the cell state, and the cell output are denoted by $\bm{x}$, $\bm{c}$ and $\bm{h}$, respectively. The cell output is then passed to a single layer perceptron with a linear activation, an addition to the standard LSTM cell architecture to keep the input and output ($\hat{\bm{y}}$) dimensions consistent.  

Here, we employ a stacked LSTM architecture and perform recursive operations to enable the sequence-to-sequence learning, as expressed with the help of the following equations:
\begin{equation} 
    \label{eq1}
    \begin{split}
        \bm{h}_k = \bm{F}\Big(\bm{c}_{k-1}, \bm{h}_{k-1}, \bm{x}_k\Big), & \quad \bm{c}_k = \bm{H}\Big(\bm{c}_{k-1},\bm{h}_{k-1}, \bm{x}_k\Big), \\
        \hat{\bm{y}}_k & = \bm{G}\Big(\bm{h}_{T_x+k}\Big) \\
        \bm{X} = \Big[\bm{x}_1, \, \bm{x}_2, \, \ldots \, , \, \bm{x}_{Tx-1}, \, \bm{x}_{Tx} \Big], & \quad \bm{\hat{Y}} = \Big[\hat{\bm{y}}_1, \, \hat{\bm{y}}_2, \, \ldots \, , \, \hat{\bm{y}}_{Tx-1}, \, \hat{\bm{y}}_{Tx} \Big], \\ 
        \bm{x}_{Tx+1} = \hat{\bm{y}}^1, \ \bm{x}_{Tx+2} = \hat{\bm{y}}^2, \, \ldots \, ,& \ \bm{x}_{2 Tx-2} = \hat{\bm{y}}^{Tx-2}, \ \bm{x}_{2Tx-1} = \hat{\bm{y}}^{Tx-1}. 
    \end{split}
\end{equation}
Here $\bm{F}$, $\bm{G}$ are the nonlinear activation functions and $\bm{H}$ is the linear activation function of the connected perceptron layer. The subscripts refer to the element number of the input sequence and the superscripts refer to the element number of the output sequence. The LSTM is run recursively for $Tx$ iterations, where $Tx$ is the length of the input sequence, to obtain the next $Tx$ sequences. For this study, $Tx$ was selected as 9, thus discretizing the temporal sequence into 9 uniform time-steps. 

\begin{figure}[ht!]
    \centering
    \includegraphics[width=0.85\linewidth]{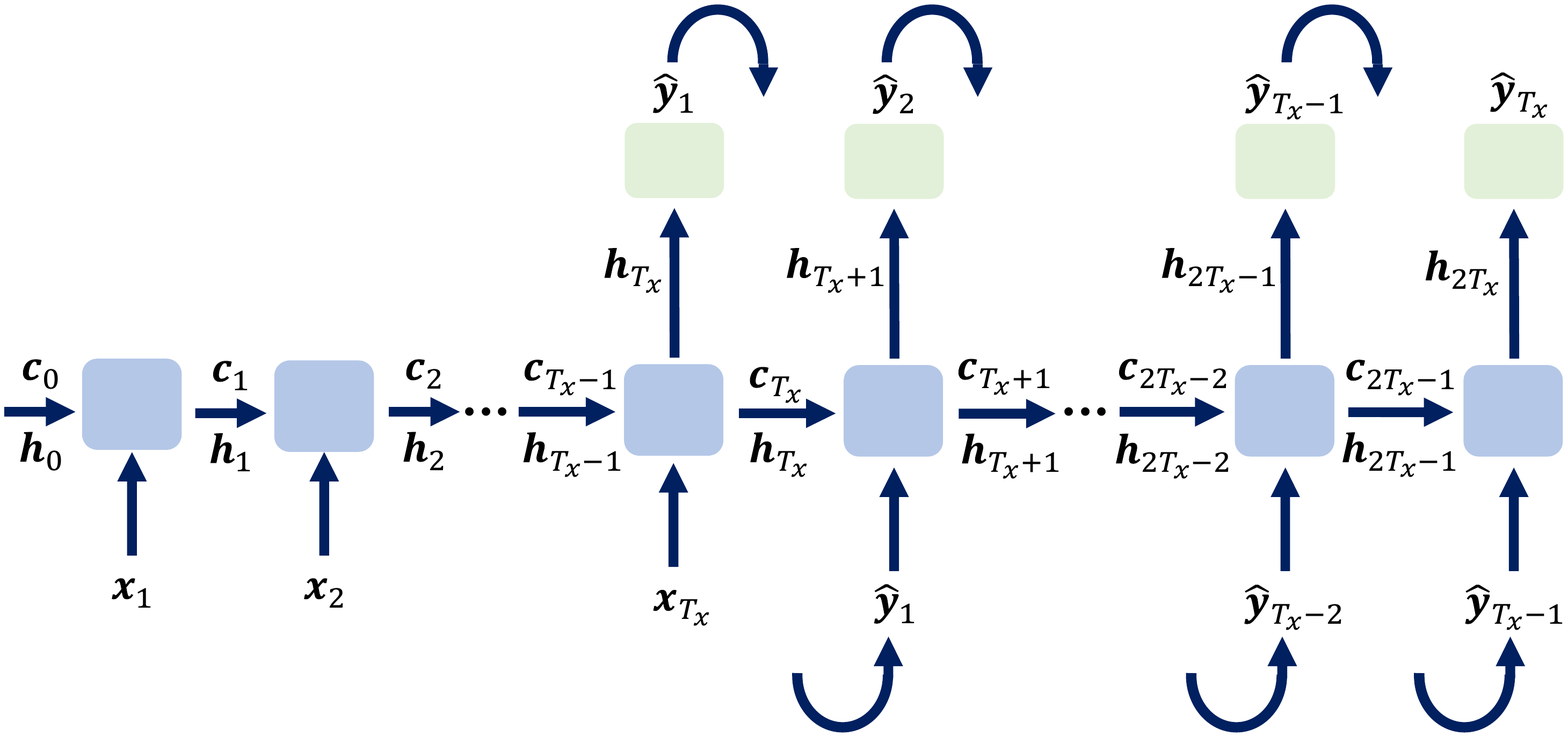}
    \caption{Recursive LSTM architecture}
    \label{fig:lstm_architecture}
\end{figure}

A schematic of the recursive LSTM architecture showing the connection from the first to the second iteration of the recursion loop is shown in Fig.~\ref{fig:lstm_architecture}. Here only the new output state is requested as output from the LSTM. These new outputs are used to form the updated input to the LSTM for the next iteration and the operation is performed recursively until a completely new output sequence of $Tx$ output states is obtained \cite{tensorflow2015-whitepaper}.

ADAM optimizer was employed for training the network with an initial learning rate of 0.0005, a learning rate decay of 0.9, and a decay step of 1000 epochs. The network is trained for 3500 epochs, which results in satisfactory convergence of the training error for the plain LSTM network. The recursive LSTM architecture was developed and trained in TensorFlow 2.3.0.

\subsubsection{Application of kinematical consistency}
The kinematical consistency is applied by augmenting a regularizer to the standard mean square error loss function of the recursive LSTM architecture. The regularizer computes a scaled mean square error between the wave speeds $c_0$ and $\hat{c}_0$, computed from the target and predicted pressure fluctuation signals, $\acute{p}(x,t)$ and $\hat{\acute{p}}(x,t)$, respectively. The modified loss functions can be written as follows,
\begin{equation}
    \label{eq:mod_loss_func}
    \mathscr{L} = \frac{1}{2} {\left( \hat{\acute{p}} - \acute{p} \right)}^2 + \lambda^2 {\left( \frac{\hat{c}_0}{c_0} - 1\right)}^2.
\end{equation}
The scaling factor, $\lambda$, is a hyperparameter of the learning problem and is dimensionally analogous to an artificial bulk modulus. The optimal value of $\lambda$ is the artificial bulk modulus required by the ML scheme to not only predict sufficiently accurate solutions but also ensure that the predictions follow the characteristics $t - t_0 \pm \frac{x - x_0}{c_0}$ as closely as possible.   

\subsubsection{Computation of wave speed}
The physical regularization proposed in this research depends on a data-driven computation of wave speed from both the training and predicted pressure fluctuation signals, $\acute{p}(x,t)$ and $\hat{\acute{p}}(x,t)$, respectively. Computation of the wave speed $c_0$ is performed by using the kinematics of the wave propagation physics as follows:
\begin{equation}
    c_0 = \left| \frac{dx}{dt} \right| = \left| \frac{\partial \acute{p}}{\partial t} \Big\slash \frac{\partial \acute{p}}{\partial x} \right|.
\end{equation}
$\hat{c}_0$ can be computed from $\hat{\acute{p}}(x,t)$ in an analogous manner. Minimizing the mean square errors between target and predicted wave speeds, even as a regularization operation, leads to a greater resemblance between predicted and real characteristic patterns, and induces the desired kinematical consistency to the LSTM network.

It is also important to note the limitation of the present data-driven wave speed computation methodology for such physical problems. The wave speed is presently computed via second-order finite difference approximation of the partial derivatives. This poses a problem at any point where two (or more) waves interfere, as the principle of superposition of the linear waves ($\acute{p} = \acute{p}_1 + \acute{p}_2$) does not apply to the problem kinematics as shown below 
\begin{equation}
    \left| \frac{dx}{dt} \right| = \left| \frac{\partial \acute{p}}{\partial t} \Big\slash \frac{\partial \acute{p}}{\partial x} \right| = \left| \Big(\frac{\partial \acute{p}_1}{\partial t} + \frac{\partial \acute{p}_2}{\partial t}\Big) \Big\slash \Big(\frac{\partial \acute{p}_1}{\partial x} + \frac{\partial \acute{p}_2}{\partial x}\Big) \right| \neq \left| \frac{\partial \acute{p}_1}{\partial t} \Big\slash \frac{\partial \acute{p}_1}{\partial x} + \frac{\partial \acute{p}_2}{\partial t} \Big\slash \frac{\partial \acute{p}_2}{\partial x} \right|.
\end{equation}
Thus, it would require separate computation of the partial derivatives of two or more unknown pressure fluctuations at the inference, which is an intractable task. This restricts the present approach to those regions of the space-time domain that are free of interference. However, as shown later in the results, even for a moderate number of interferences the present approach for the wave speed computation is still applicable to a significant part of the space-time domain.

\section{Results and discussion}
\subsection{Computation of wave speed}
\begin{figure}[ht]
    \centering
    \begin{subfigure}{.32\textwidth}
        \centering
        \includegraphics[width=\linewidth]{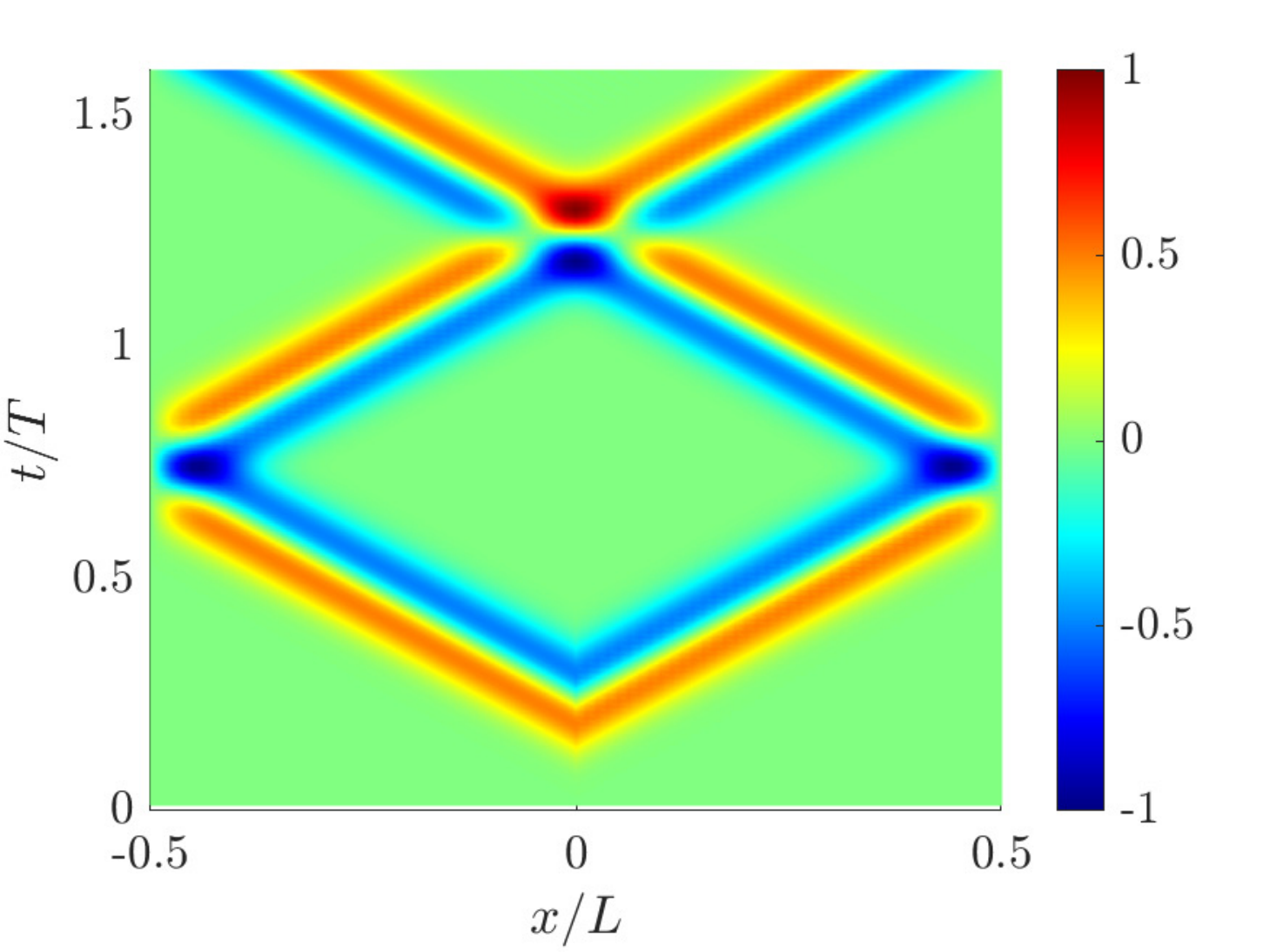}
        \caption{Pressure fluctuation training data}
    \end{subfigure}
    \hfill
    \begin{subfigure}{.32\textwidth}
        \centering
        \includegraphics[width=\linewidth]{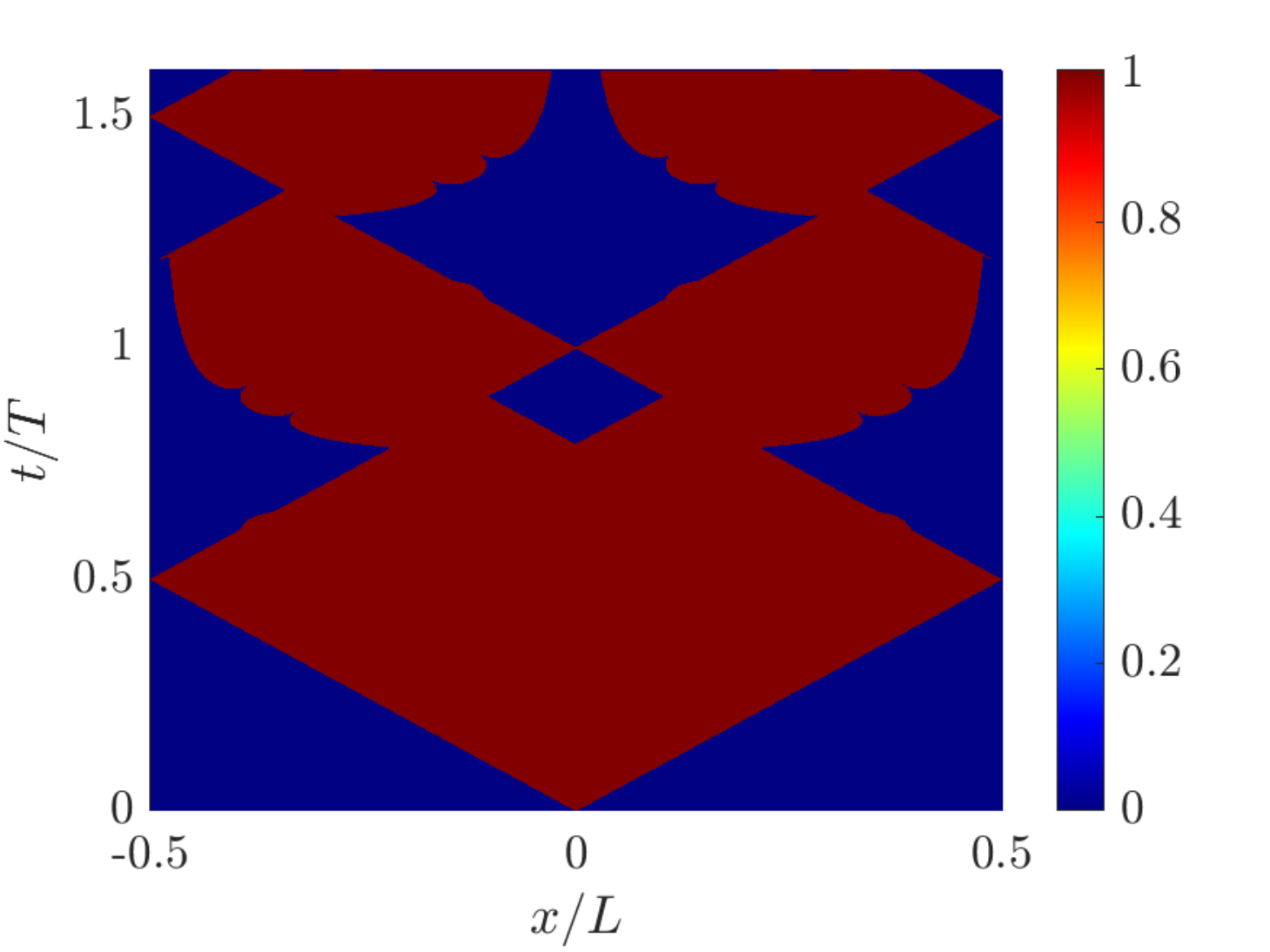}
        \caption{Wave speed: FDM mesh}
    \end{subfigure}
    \hfill
    \begin{subfigure}{.32\textwidth}
        \centering
        \includegraphics[width=\linewidth]{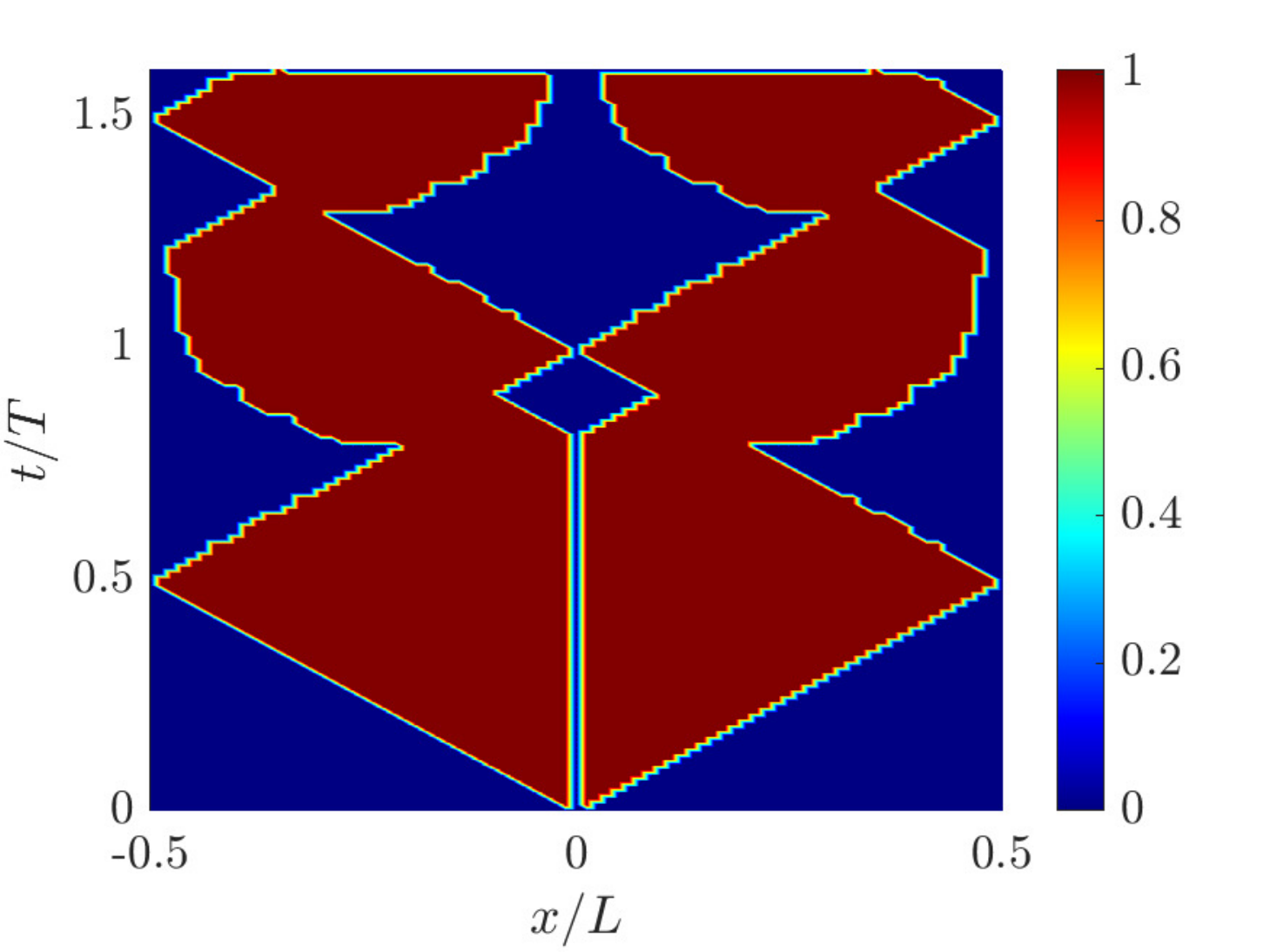}
        \caption{Wave speed: ML mesh}
    \end{subfigure}
    \caption{Wave speed computation}
    \label{fig:wavespeed_case1}
\end{figure}

The training data used for the learning problem was generated by a finite difference scheme with a second-order accurate centered difference in space and an explicit forward difference in time. A Gaussian monopole source was applied at the center of the domain with a delay of $t_0=0.25T$, where $T$ is the time period of the wave characteristics. The training data shown in Fig.~\ref{fig:wavespeed_case1} (a), consists of only 13 examples of input sequences of length $T_x=0.1143T$ . The training data contains a single instance of reflection at both the right and left walls, and a single interference event of the right and left traveling waves. 

The wave speed computed on the training data obtained with a very fine mesh is shown in Fig.~\ref{fig:wavespeed_case1} (b). As explained earlier, the computation predicted the exact wave speed at all locations not undergoing any interference. At those regions where wave interference was detected from the data, the wave speed was set to zero to ensure that they do not participate in the LSTM regularization process. The other regions on the space-time domain that show zero wave speed are regions so far from the characteristic lines that they do not observe any detectable pressure fluctuation. The training data actually provided to the network was first interpolated to a 16 times coarser mesh for both spatial and temporal discretization. However, as demonstrated by the wave speed distribution on the 16 times coarser space-time discretization (Fig.~\ref{fig:wavespeed_case1} (c)) compared to the much finer mesh (Fig.~\ref{fig:wavespeed_case1} (b)), the largest scales involved in the problem kinematics are not affected by the coarse-graining. However, this coarse-graining significantly improves the training efficiency and accuracy as learning is performed on a smaller feature space with fewer gradient evaluations and smaller error accruing \cite{bronstein2021geometric}.

\subsection{Impact on LSTM forecasting capability}
\begin{figure}[ht]
    \centering
    \includegraphics[width=.7\linewidth]{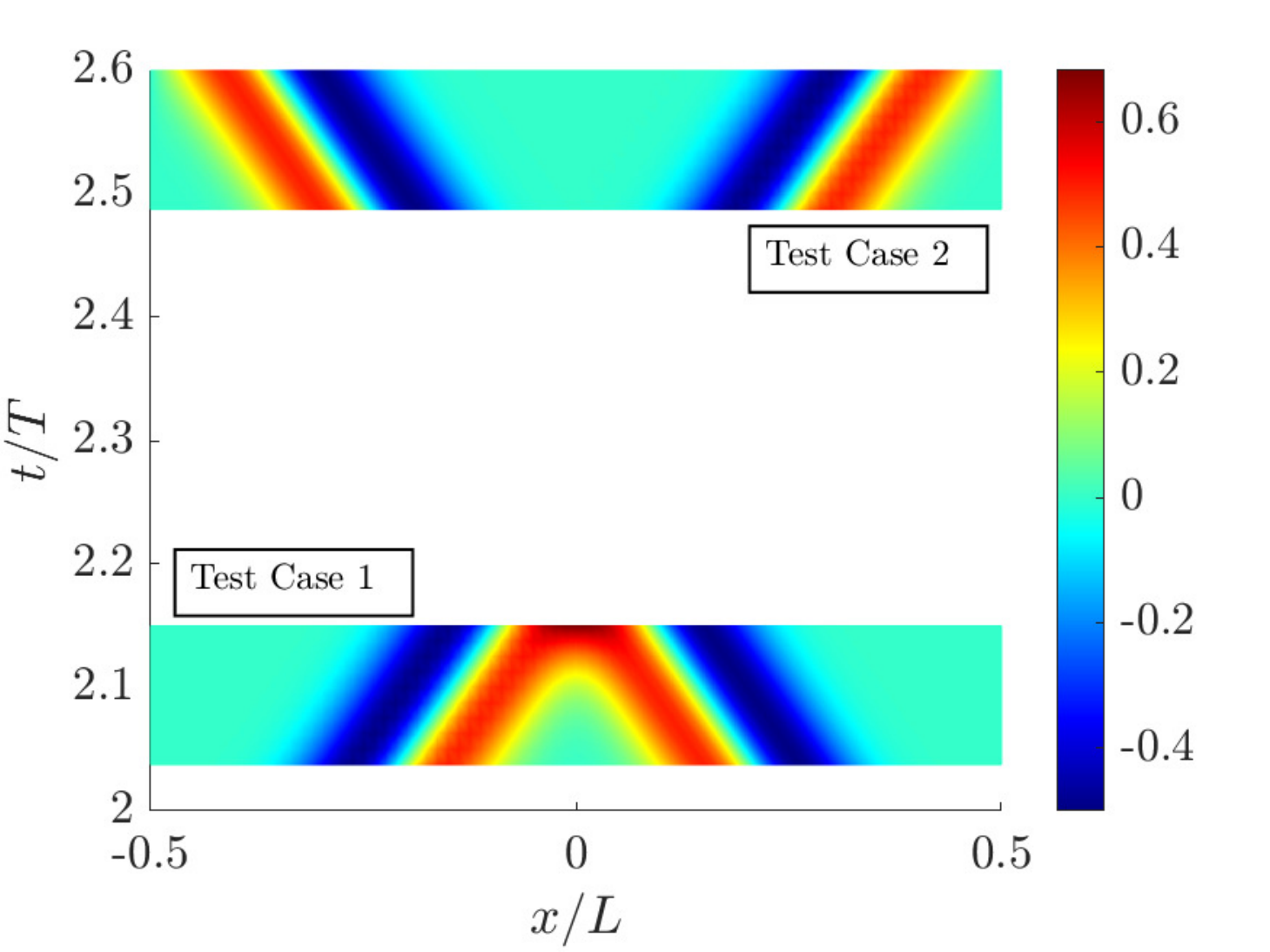}
    \caption{Test inputs}
    \label{fig:test_data}
\end{figure}

We first study the impact of the kinematical consistency on the forecasting capability of the physics-based LSTM network compared to that of the plain LSTM network. The input sequences to the LSTM network for the two test cases considered are shown in Fig.~\ref{fig:test_data}. Test Case 1 considers a case where we intend to forecast the wave propagation after it undergoes interference at the center of the domain and Test Case 2 considers forecasting after reflection from the boundary. These types of events are considered difficult forecasting cases for wave propagation phenomena \cite{sorteberg2019approximating} and so we will particularly focus on them.

\begin{figure}[ht]
    \centering
    \begin{subfigure}{.245\textwidth}
        \centering
        \includegraphics[width=\linewidth]{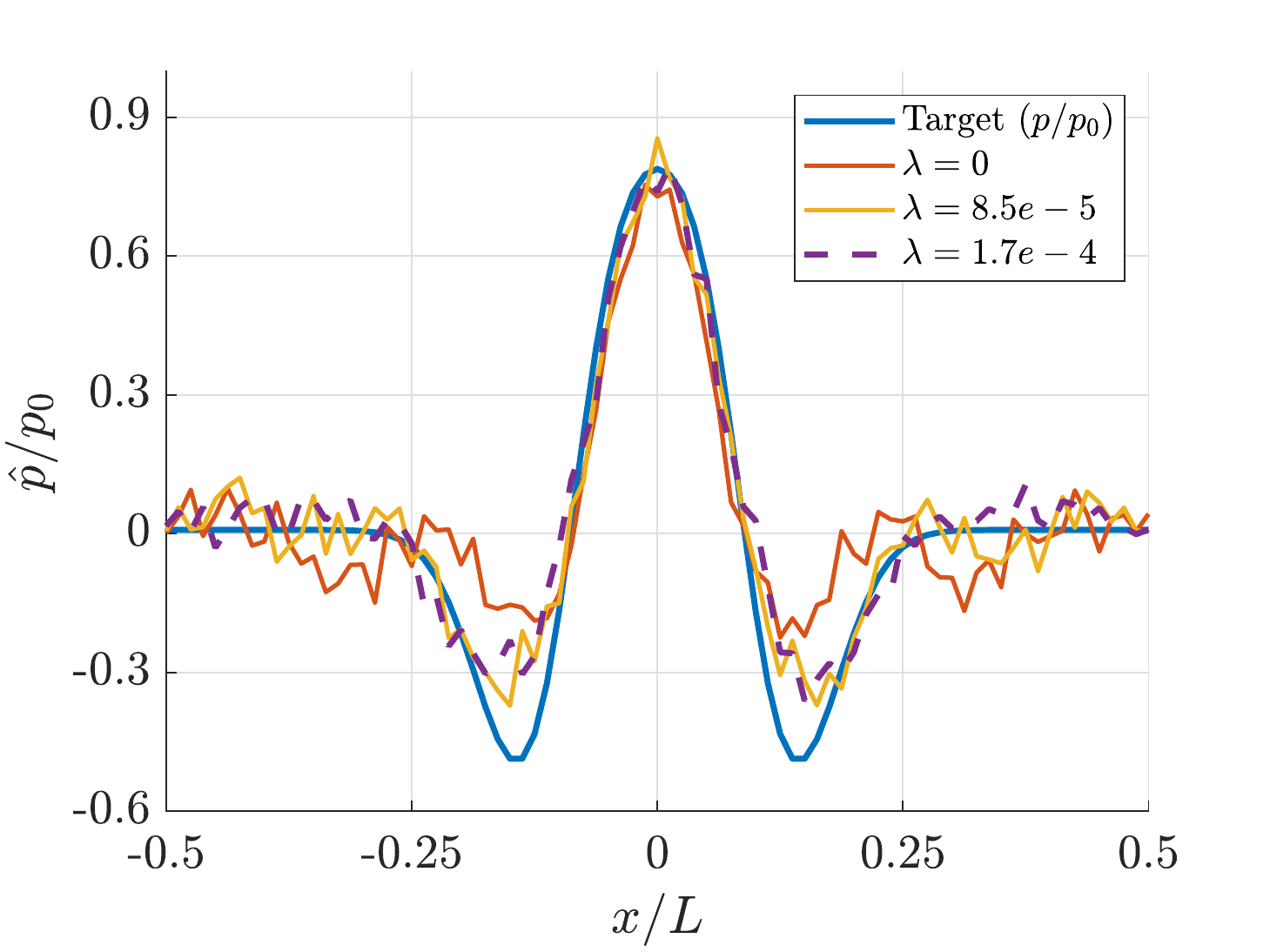}
        \caption{$t=0.0381T$}
    \end{subfigure}
    \begin{subfigure}{.245\textwidth}
        \centering
        \includegraphics[width=\linewidth]{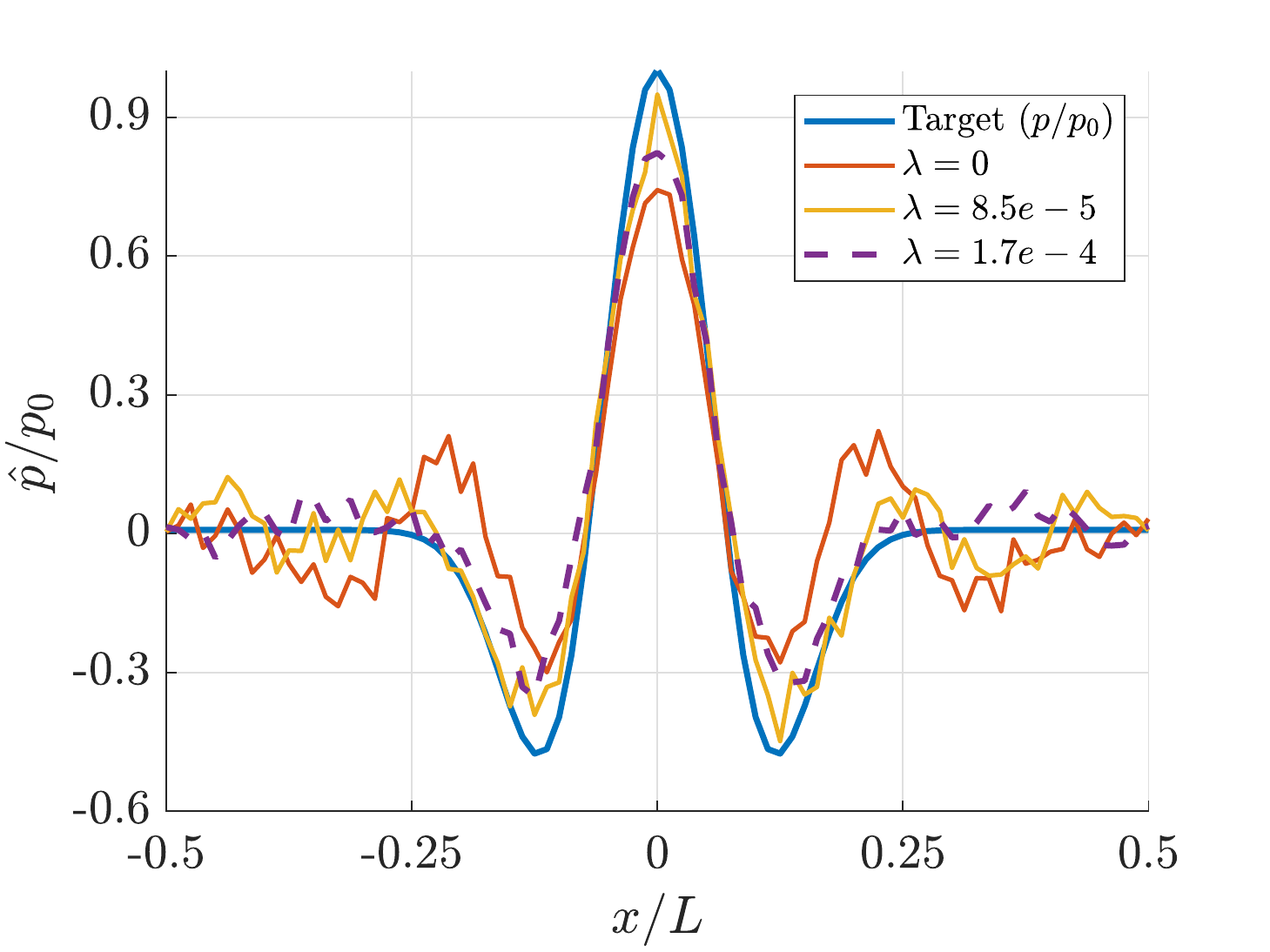}
        \caption{$t=0.0635T$}
    \end{subfigure}
    \begin{subfigure}{.245\textwidth}
        \centering
        \includegraphics[width=\linewidth]{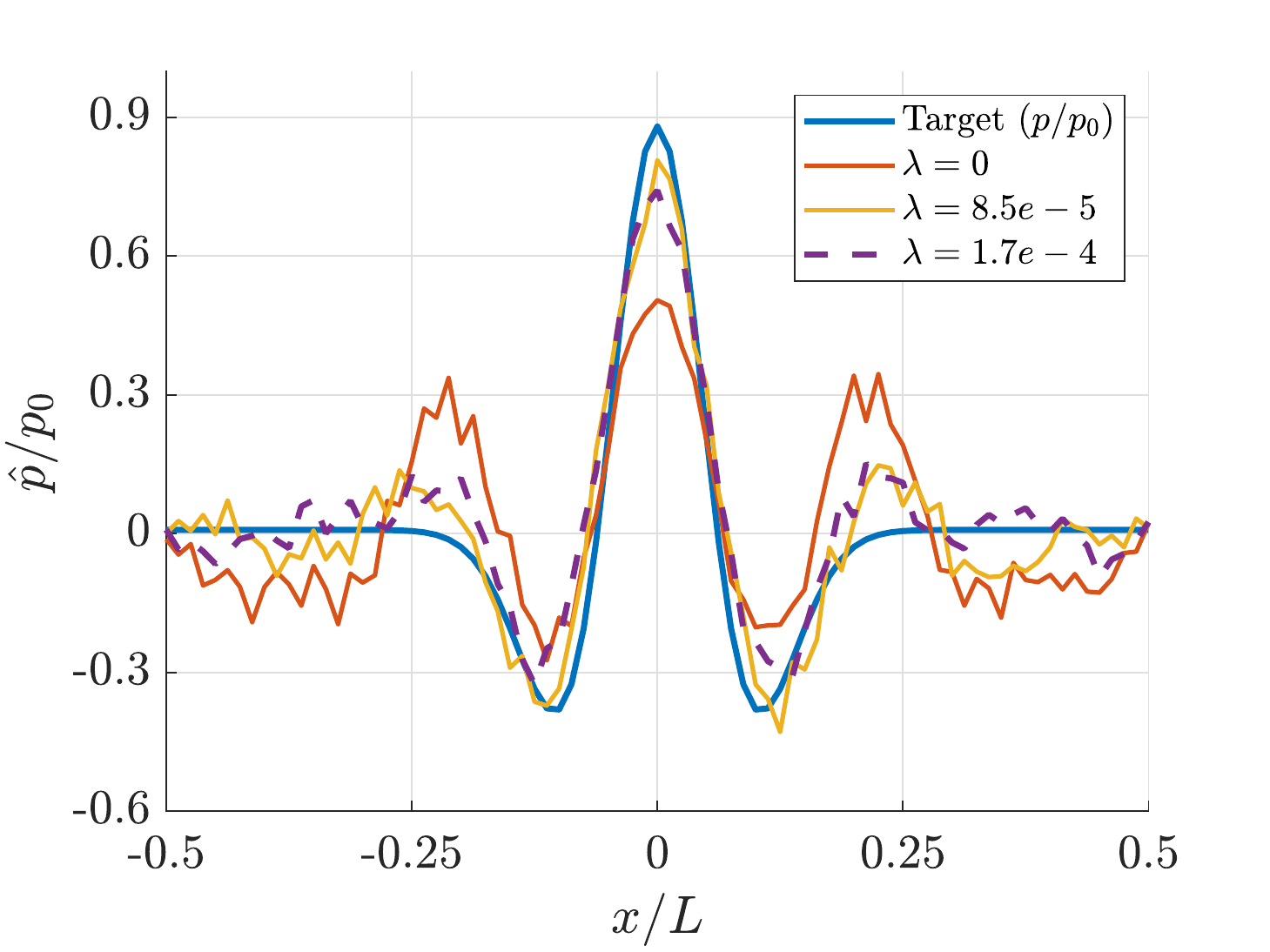}
        \caption{$t=0.0889T$}
    \end{subfigure}
    \begin{subfigure}{.245\textwidth}
        \centering
        \includegraphics[width=\linewidth]{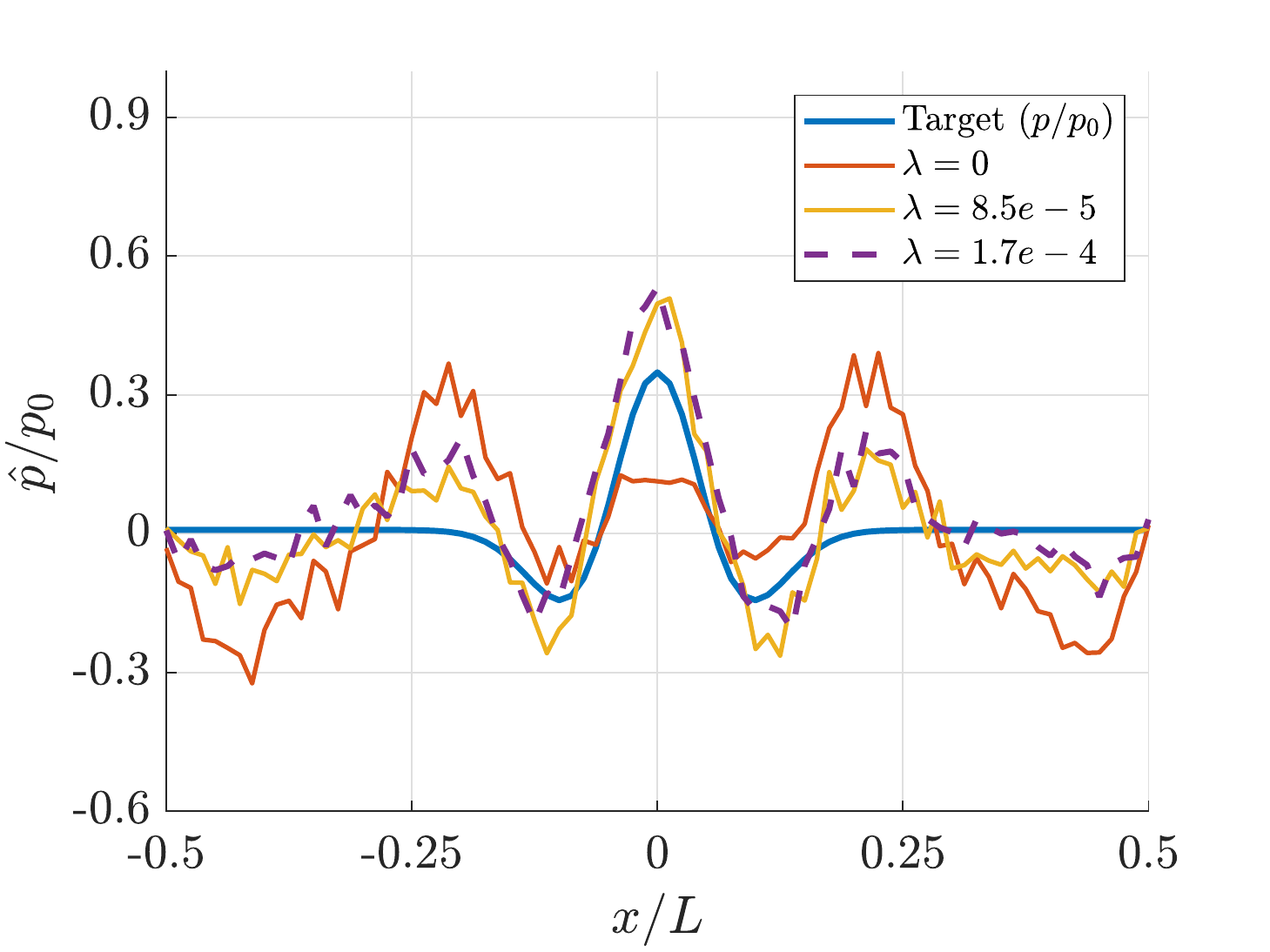}
        \caption{$t=0.1143T$}
    \end{subfigure}
    \caption{Comparison of test predictions: Case 1}
    \label{fig:sol_comp_samp5}
\end{figure}

The normalized predicted signals for Test Case 1 are shown in Fig~\ref{fig:sol_comp_samp5} (a), (b), (c) and (d), for $t=0.0381T$, $t=0.0635T$, $t=0.0889T$ and $t=0.1143T$ into the horizon, respectively. Here, we denote the target and predicted signals as $\bm{p}$ and $\hat{\bm{p}}$, respectively, and $p_0$ is the peak pressure amplitude of the training set. The normalized pressure difference is denoted as $\bm{e} = \frac{\hat{\bm{p}}}{\bm{p}} - 1$. The results in Fig.~\ref{fig:sol_comp_samp5} show that the plain LSTM ($\lambda=0$) not only fails to predict the peak amplitude accurately beyond $t=0.0635T$ but severe inaccuracies are observed at various spatial locations for all the signals. On the other hand, the physics-based LSTM network with $\lambda=8.5\times 10^{-5}$ and $\lambda=1.7\times 10^{-4}$ provide much better predictions for both the peak amplitude as well as the low-frequency oscillation amplitudes near $x/L=\pm 0.25$. The only inaccuracies are observed near the peak for $t=0.1143T$ but with a smaller error than the plain LSTM.

\begin{figure}[ht]
    \centering
    \begin{subfigure}{.245\textwidth}
        \centering
        \includegraphics[width=\linewidth]{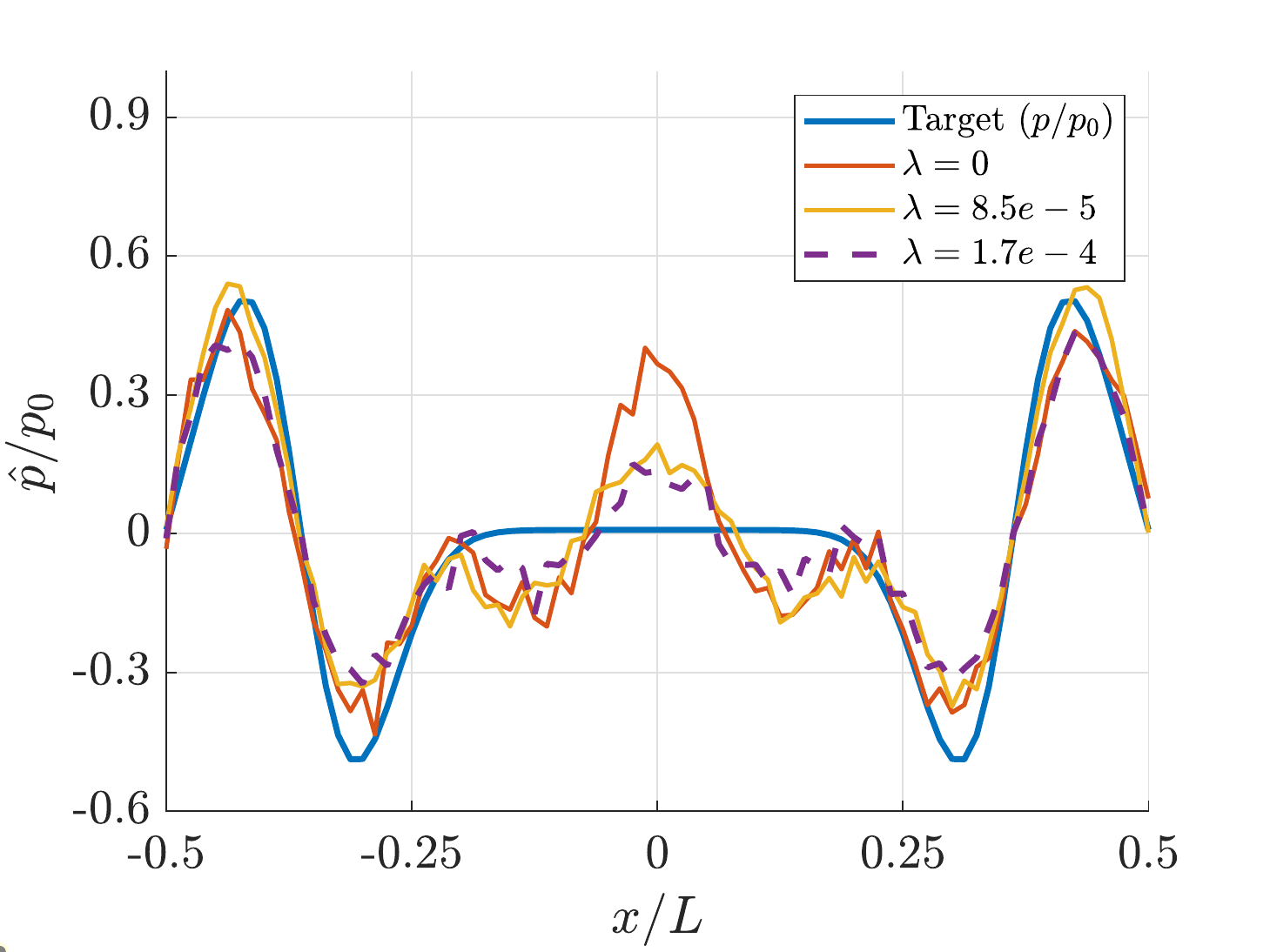}
        \caption{$t=0.0381T$}
    \end{subfigure}
    \begin{subfigure}{.245\textwidth}
        \centering
        \includegraphics[width=\linewidth]{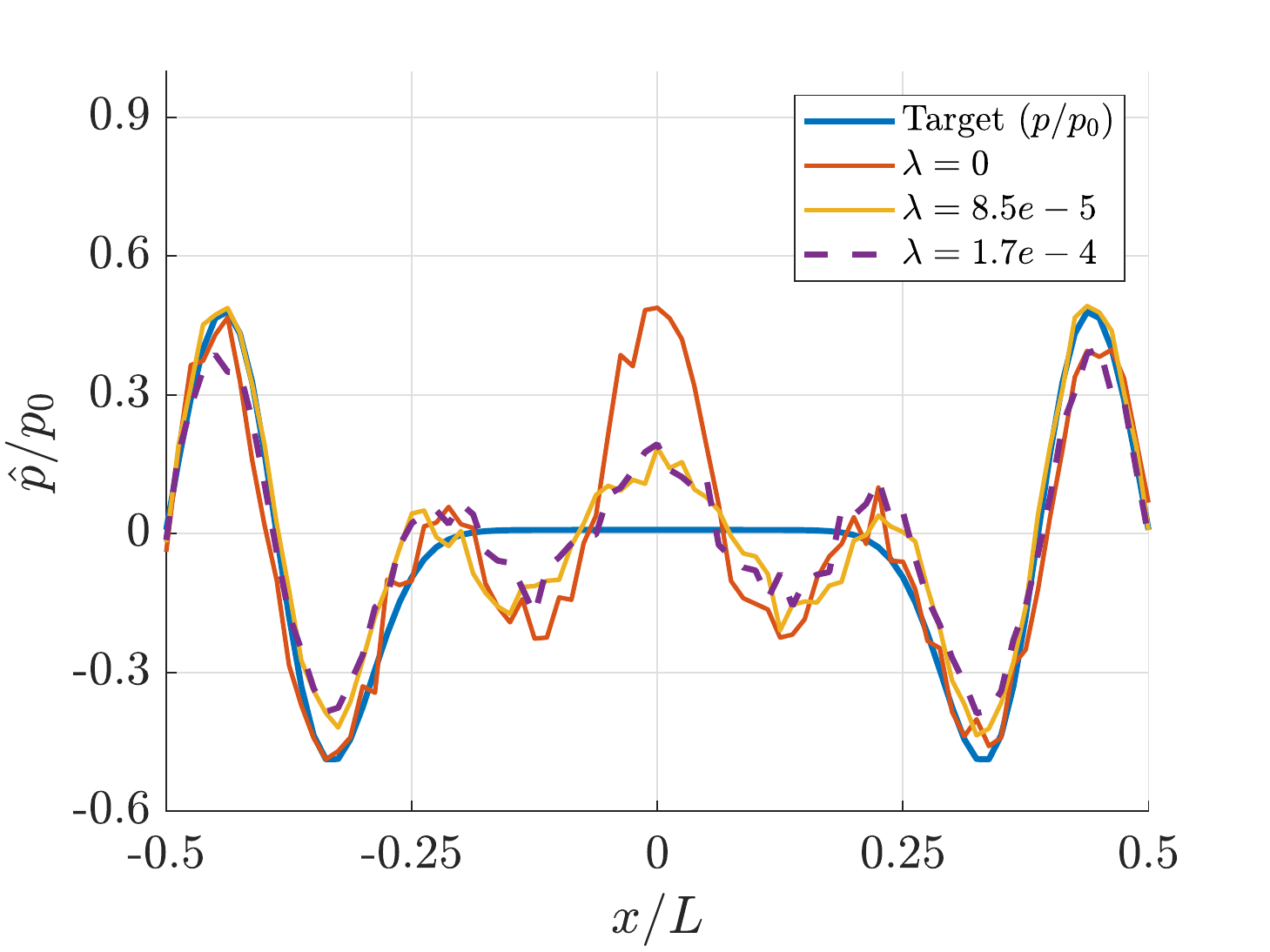}
        \caption{$t=0.0635T$}
    \end{subfigure}
    \begin{subfigure}{.245\textwidth}
        \centering
        \includegraphics[width=\linewidth]{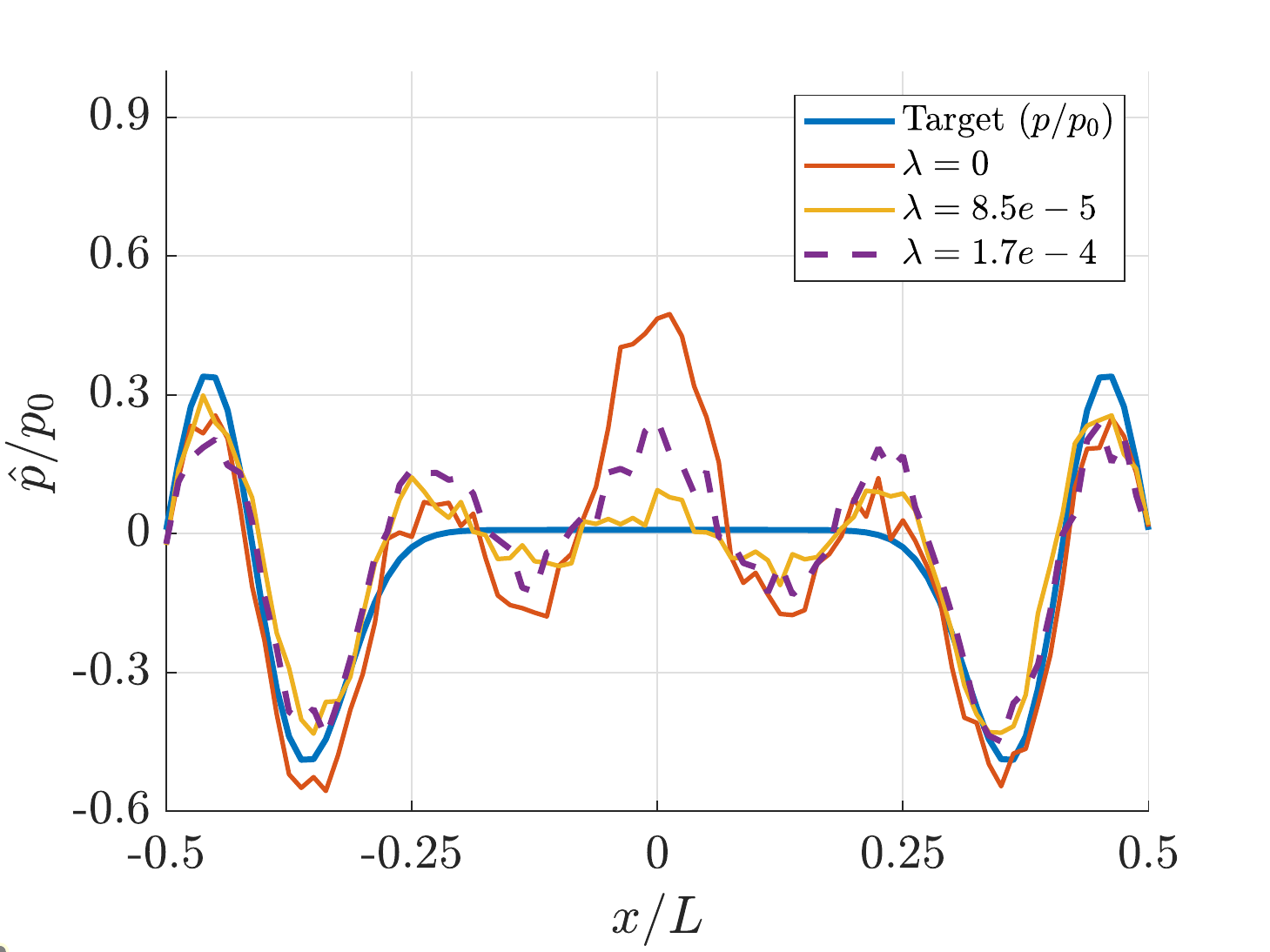}
        \caption{$t=0.0889T$}
    \end{subfigure}
    \begin{subfigure}{.245\textwidth}
        \centering
        \includegraphics[width=\linewidth]{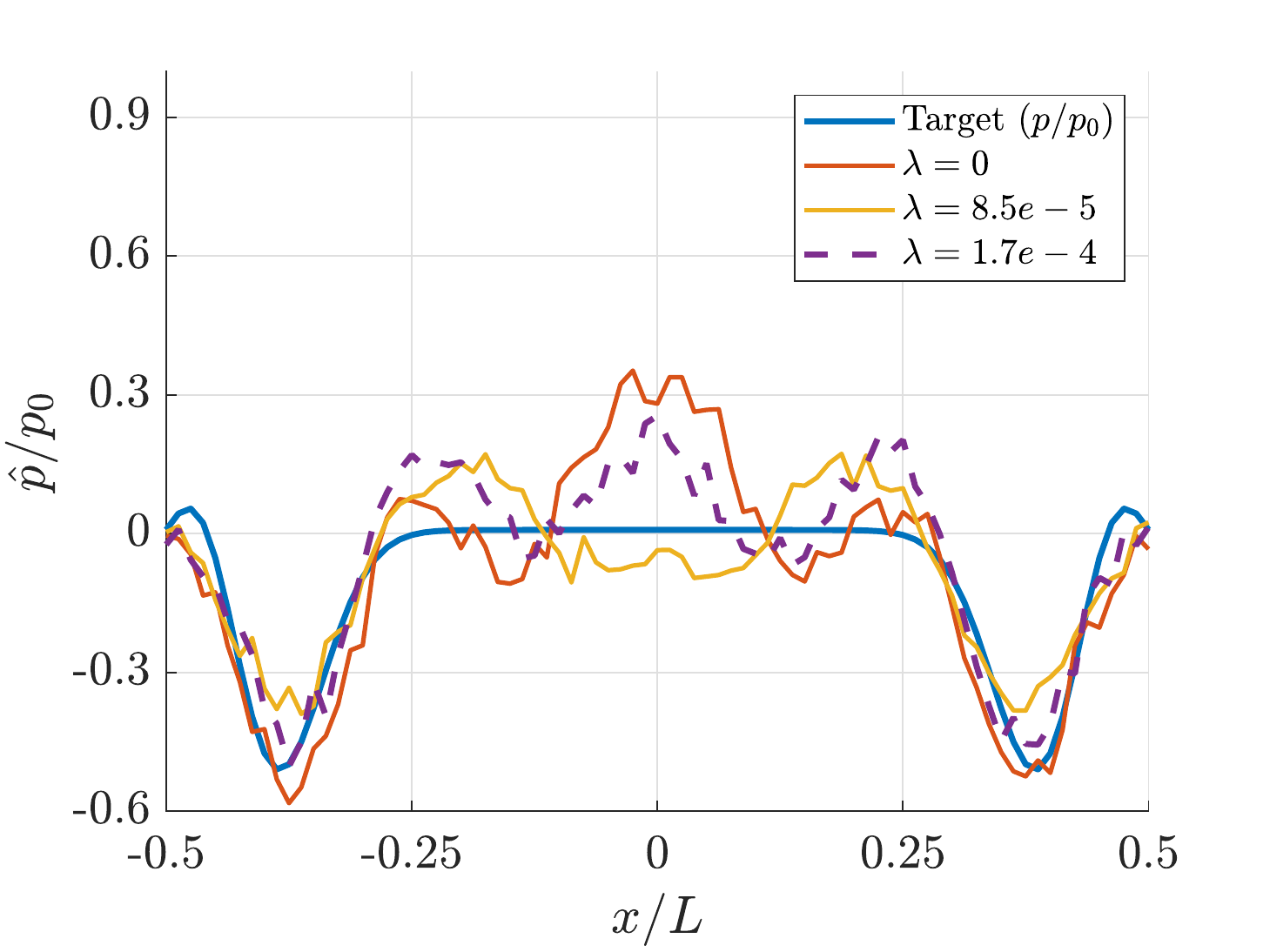}
        \caption{$t=0.1143T$}
    \end{subfigure}
    \caption{Comparison of test predictions: Case 2}
    \label{fig:sol_comp_samp9}
\end{figure}

The predicted signals for Test Case 2 are shown in Fig.~\ref{fig:sol_comp_samp9} (a), (b), (c), and (d), for for $t=0.0381T$, $t=0.0635T$, $t=0.0889T$ and $t=0.1143T$ into the horizon, respectively. The results show that although the overall forecasting of the plain LSTM for this test case is better than Test Case 1, yet significantly large inaccuracies are observed near $x/L=0$. These errors increase as we go further into the horizon until it reaches $t=0.1143T$. For $t=0.1143T$, although the predictions improve near the center of the domain, fluctuating inaccuracies can be observed in other parts of the spatial domain. The physics-based LSTMs are able to significantly reduce the largely concentrated errors of the plain LSTM around $x/L=0$ without adversely affecting other parts of the spatial domain.

\begin{figure}[ht]
    \centering
    \begin{subfigure}{.49\textwidth}
        \centering
        \includegraphics[width=\linewidth]{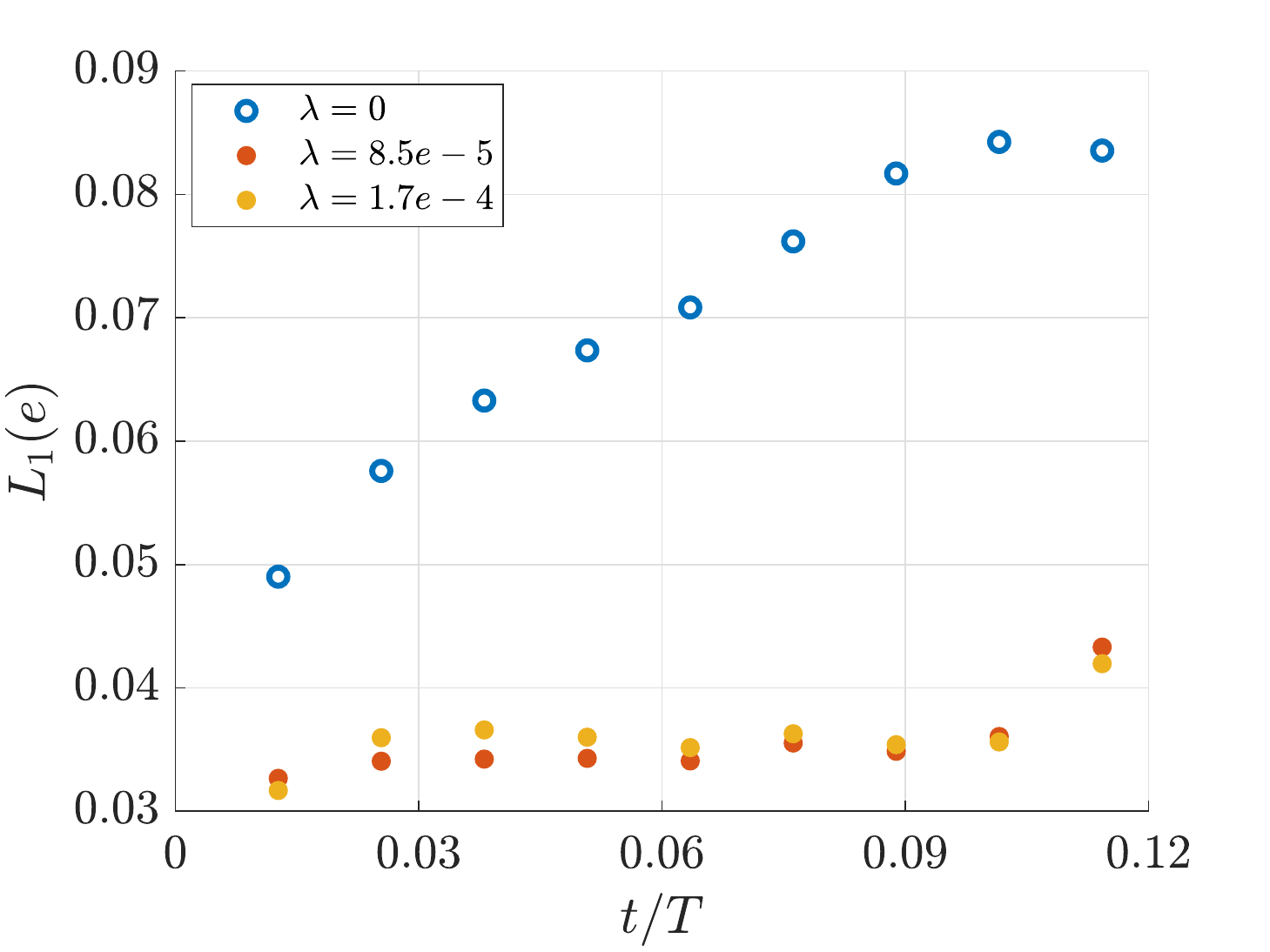}
        \caption{$L_1$ error norm}
    \end{subfigure}
    \hfill
    \begin{subfigure}{.49\textwidth}
        \centering
        \includegraphics[width=\linewidth]{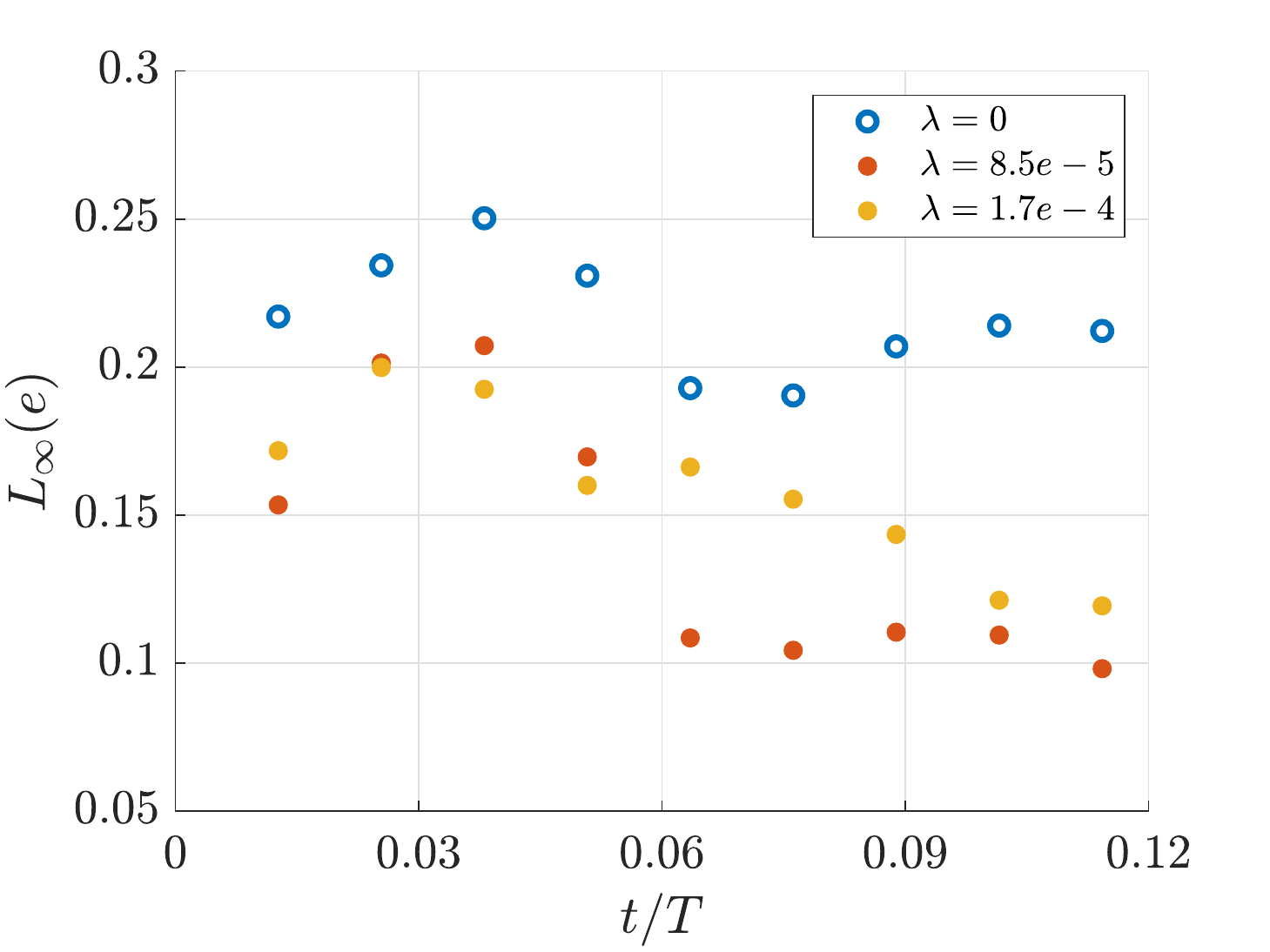}
        \caption{$L_\infty$ error norm}
    \end{subfigure}
    \caption{Comparison of prediction errors: Case 1}
    \label{fig:pred_errs_case1}
\end{figure}

\begin{figure}[ht]
    \centering
    \begin{subfigure}{.495\textwidth}
        \centering
        \includegraphics[width=\linewidth]{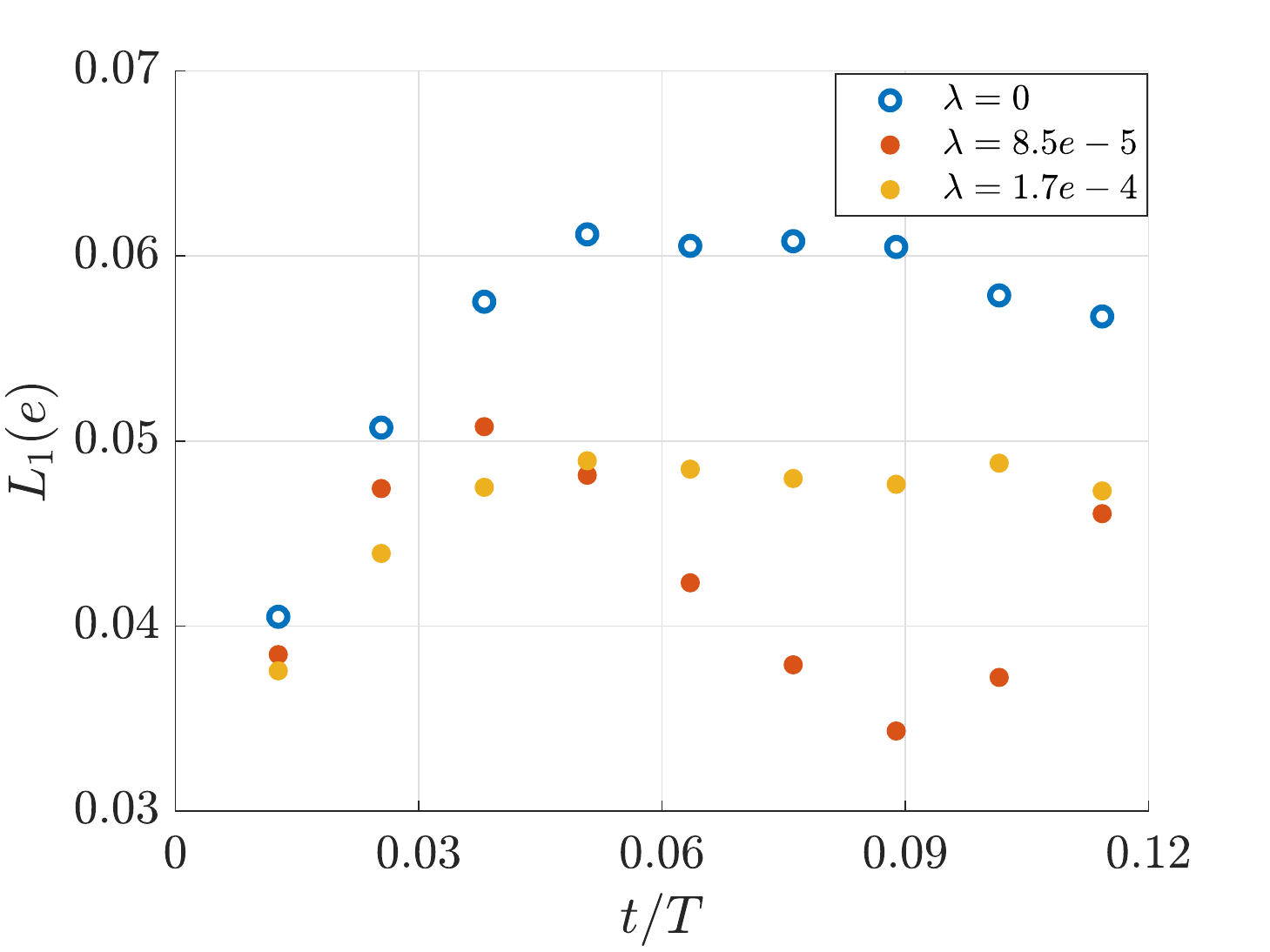}
        \caption{$L_1$ error norm}
    \end{subfigure}
    \hfill
    \begin{subfigure}{.495\textwidth}
        \centering
        \includegraphics[width=\linewidth]{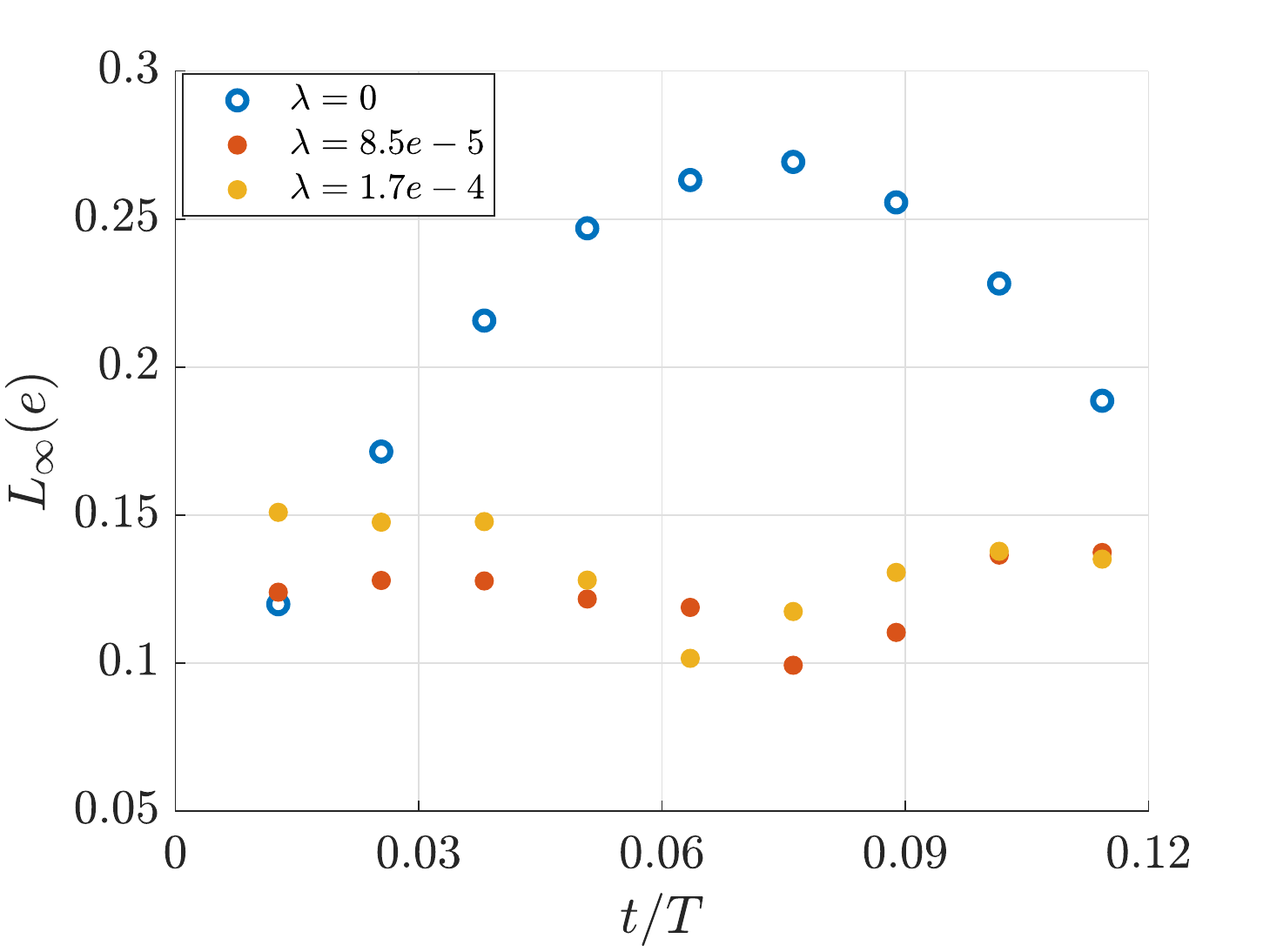}
        \caption{$L_\infty$ error norm}
    \end{subfigure}
    \caption{Comparison of prediction errors: Case 2}
    \label{fig:pred_errs_case2}
\end{figure}

The $L_1$ and $L_\infty$ error norm of the plain and physics-based LSTM predictions for Test Case 1 are compared in Fig~\ref{fig:pred_errs_case1} (a) and (b), respectively. While the $L_1$ error shows the overall behavior of the predicted results, the $L_\infty$ error can indicate the worst-case errors due to the shifts in the peak amplitudes as a result of the inaccurate wave speed of the solutions. Thus, both these error norms are presented here. The physics-based LSTM models for both the values of $\lambda$ provide a significantly smaller $L_1$ error compared to the plain LSTM. However, for $\lambda=1.7\times {10}^{-4}$, the physics-based LSTM model predictions show the smallest $L_\infty$ error norm on several cases. Similarly, we present the $L_1$ and $L_\infty$ error norms for Test Case 2 in Fig.~\ref{fig:pred_errs_case2} (a) and (b), respectively. The $L_1$ error of the plain LSTM is not as large as in Test Case 1 and decreases after $t=0.0635T$. However, they are still much larger than the errors computed from the physics-based LSTM predictions. The $L_\infty$ error norms for Test Case 2 are actually larger than those for Test Case 1, especially from $t=0.0508T$ to $t=0.0889T$. For these time steps of the sequence, large erroneous peaks were observed around $x=0$ (Fig.~\ref{fig:sol_comp_samp9} (b) and (c)) leading to the very large $L_\infty$ error norms.

\begin{figure}[ht]
    \centering
    \begin{subfigure}{.49\textwidth}
        \centering
        \includegraphics[width=\linewidth]{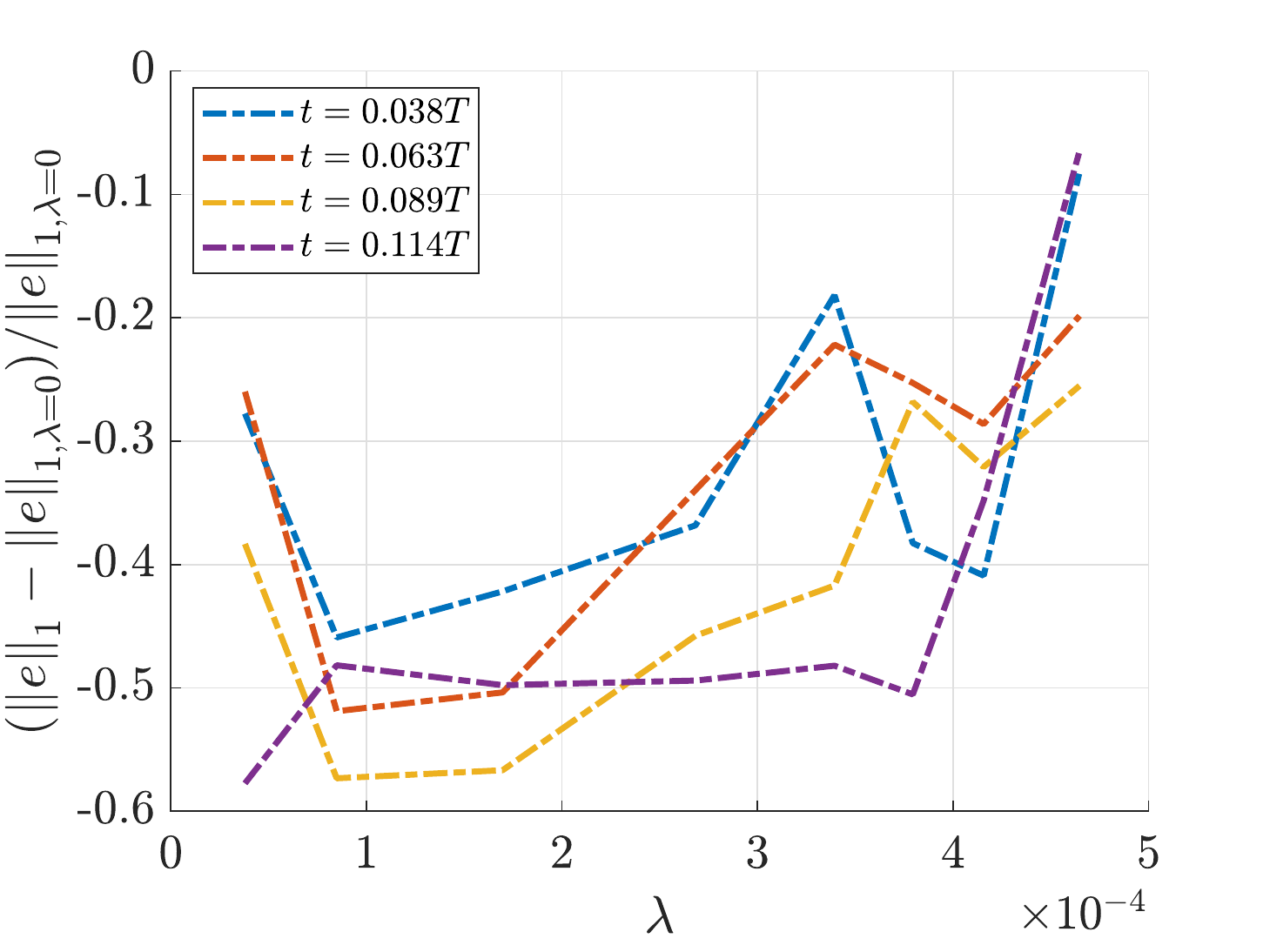}
        \caption{Relative $L_1$ error}
    \end{subfigure}
    \hfill
    \begin{subfigure}{.49\textwidth}
        \centering
        \includegraphics[width=\linewidth]{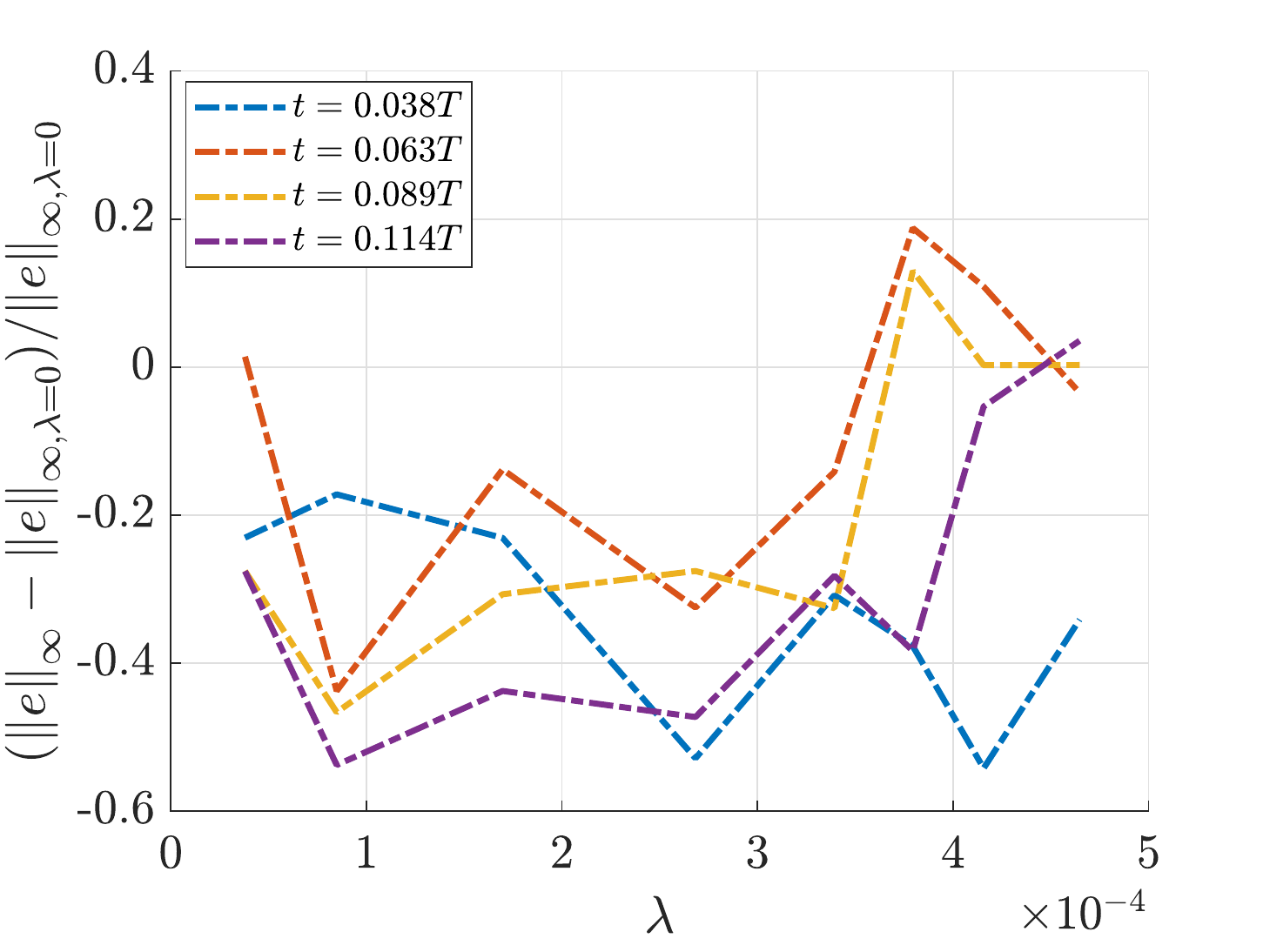}
        \caption{Relative $L_\infty$ error}
    \end{subfigure}
    \caption{Effect of $\lambda$ on relative prediction errors: Case 1}
    \label{fig:rel_pred_errs_case1}
\end{figure}

The relative difference in the $L_1$ and the $L_\infty$ error norms of the physics-based LSTM for Test Case 1 compared to its plain counterpart, and normalized by the respective plain LSTM error, is shown in Fig.~\ref{fig:rel_pred_errs_case1}. The results show that there exists an optimal range of $\lambda=8.5\times {10}^{-5}$ to $\lambda=2.3\times {10}^{-4}$ where almost a 45\% reduction in the $L_1$ error of the plain LSTM predictions is possible. However, the $L_1$ error reduction increases as we proceed further into the horizon and reaches almost 50\% for $t=0.1143T$ along with a much broader optimal range. For the $L_\infty$ error norm, the optimal range considering all the signals lies between $\lambda=7.2\times {10}^{-5}$ to $\lambda=1.7\times {10}^{-4}$, and a 20\% reduction in the $L_\infty$ error norm is possible. However, analogous to the $L_1$, the relative $L_\infty$ error reduces as we proceed further into the horizon and a 50\% reduction is possible for $t=0.1143T$ with a much broader optimal range. The plain LSTM can be expected to predict accurately only up to $t=0.0635T$ into the horizon based on Fig.~\ref{fig:sol_comp_samp5}. However, the greater accuracy of the physics-based LSTMs, especially for sequences further into the horizon, enables it to accurately predict up to $t=0.1143T$ into the horizon, which is almost twice that of the plain LSTM.

\begin{figure}[ht]
    \centering
    \begin{subfigure}{.495\textwidth}
        \centering
        \includegraphics[width=\linewidth]{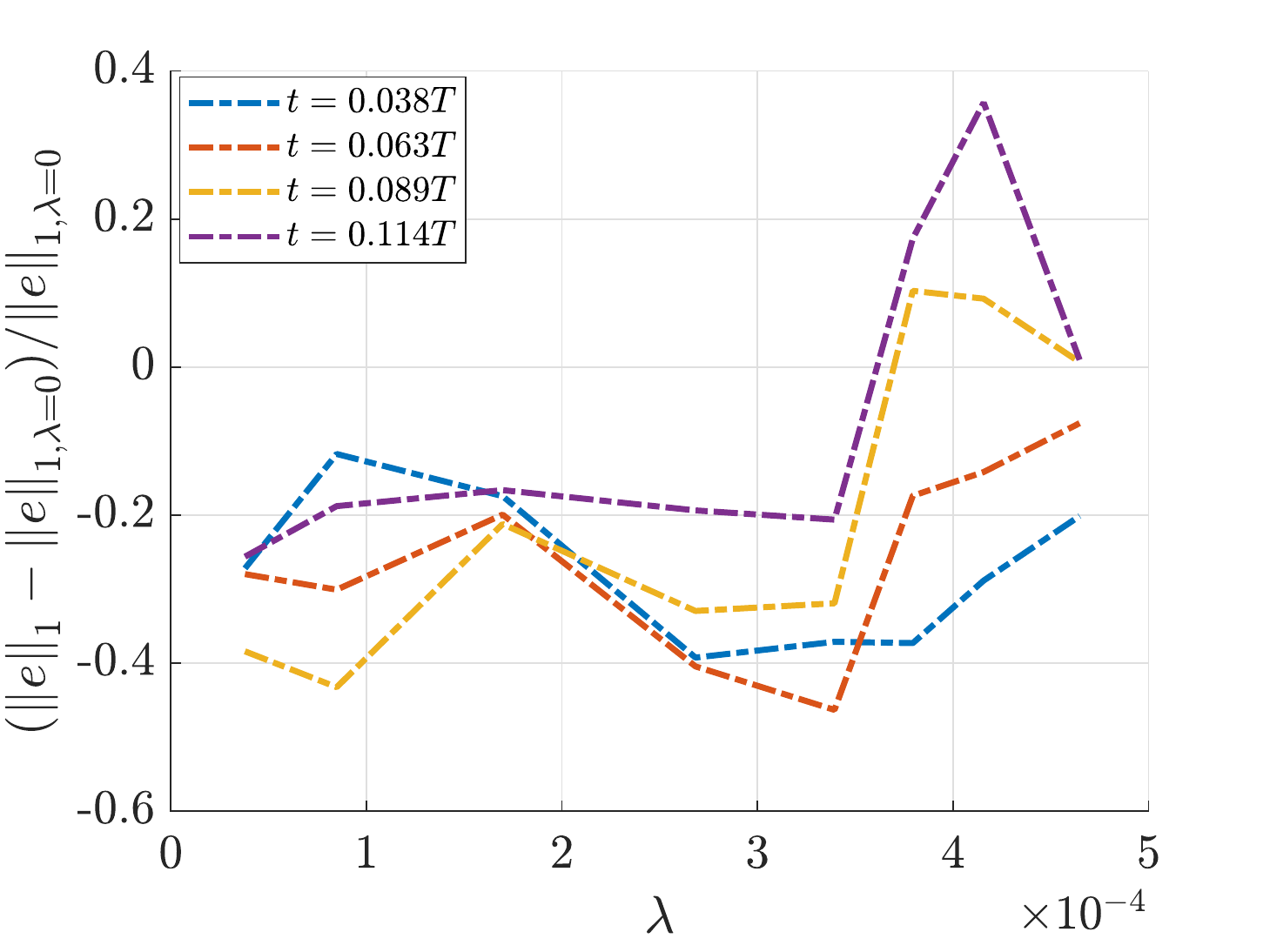}
        \caption{Relative $L_1$ error}
    \end{subfigure}
    \hfill
    \begin{subfigure}{.495\textwidth}
        \centering
        \includegraphics[width=\linewidth]{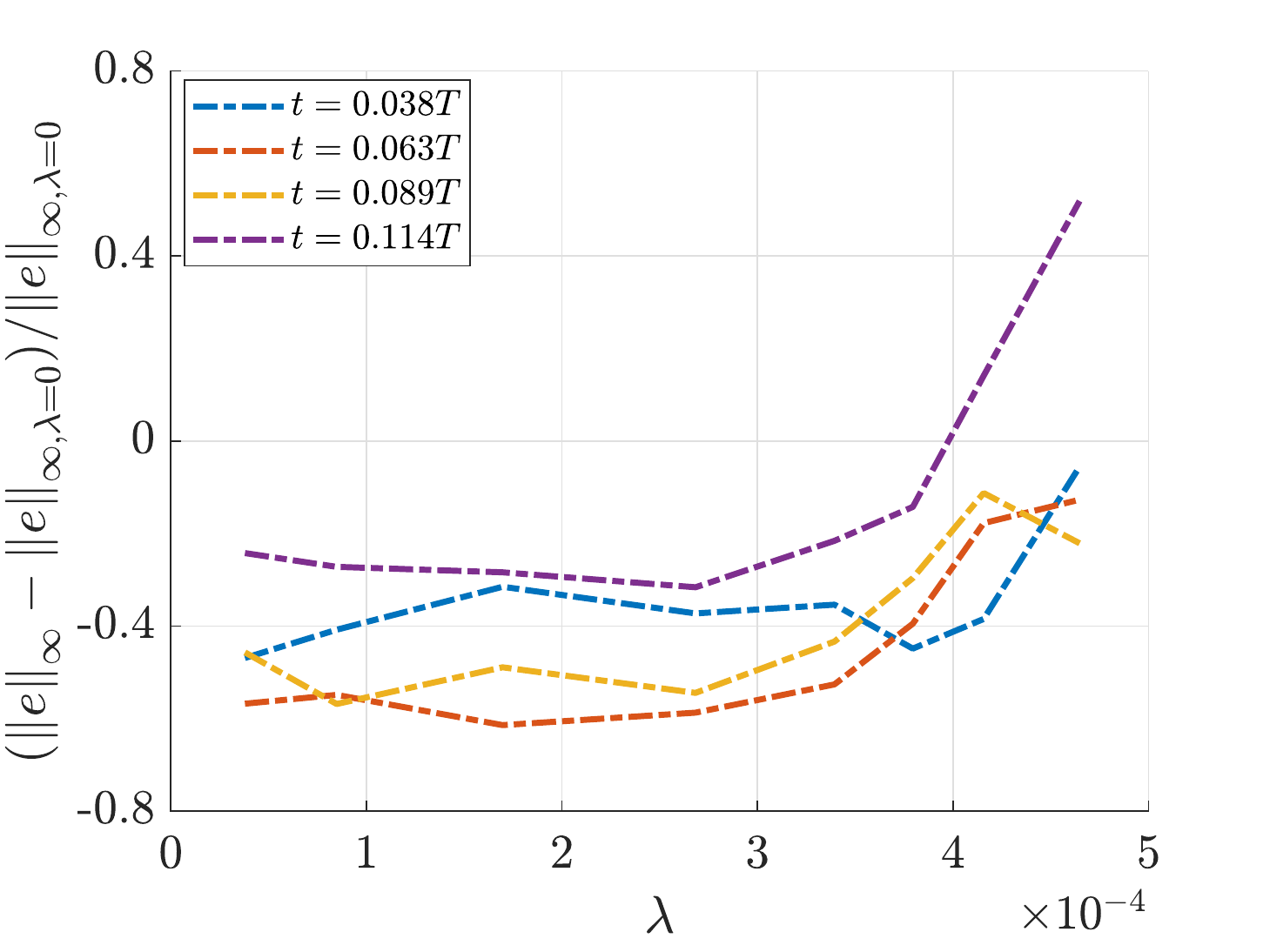}
        \caption{Relative $L_\infty$ error}
    \end{subfigure}
    \caption{Effect of $\lambda$ on relative prediction errors: Case 2}
    \label{fig:rel_pred_errs_case2}
\end{figure}

The relative $L_1$ and $L_\infty$ error norms for Test Case 2 are presented in Figs.~\ref{fig:rel_pred_errs_case2} (a) and (b), respectively. Here we see an opposite trend to Test Case 1 as both the reduction in $L_1$ and $L_\infty$ error norms are highest for the initial time steps into the horizon and lowest for the ninth time step into the horizon. Considering both the error norms for all the time steps, we can obtain an optimal range of the regularizer hyperparameter as $\lambda=3.8\times {10}^{-5}$ to $\lambda=1.7\times {10}^{-4}$. The best results overall are obtained for $\lambda=8.5\times {10}^{-5}$. Overall, using the physics-based LSTMs we can reduce the $L_1$ and $L_\infty$ error norms of the most critically inaccurate plain LSTM predictions by about 45\% and 55\%, respectively.

The results presented so far were obtained by using the best possible LSTM networks available for each value of $\lambda$ (including $\lambda=0$) from several training iterations. This became necessary as the random initial weights and biases and the random batch selection during Adam's optimization algorithms led to non-negligible variations in the test predictions. Amongst several iterations, the best network considered was the one that provided the best $L_1$ and $L_\infty$ error norms overall for both the test cases. To demonstrate this, we have provided the mean and the standard deviation of the $L_1$ and the $L_\infty$ error norms for the plain LSTM and the two optimal physics-based LSTMs most commonly used so far in Tables~\ref{tab:mean_dev_L1err_norms_TC1} and \ref{tab:mean_dev_L1inferr_norms_TC1}, respectively, for Test Case 1. Test Case 2 had similar behavior and was left out for brevity.
\begin{table}
    \caption{Mean and deviation of $L_1$ error norm: Test Case 1}
    \centering
    \begin{tabular}{cccc}
    \hline
    $t/T$ & $\lambda=0$ & $\lambda=8.5\times {10}^{-5}$ & $\lambda=1.7\times {10}^{-4}$\\\hline
    0.0127 & 0.040 $\pm$ 0.017 & 0.032 $\pm$ 0.003 & 0.038 $\pm$ 0.007\\
    0.0381 & 0.058 $\pm$ 0.035 & 0.041 $\pm$ 0.009 & 0.056 $\pm$ 0.017\\
    0.0635 & 0.072 $\pm$ 0.047 & 0.050 $\pm$ 0.018 & 0.067 $\pm$ 0.024\\
    0.0889 & 0.078 $\pm$ 0.049 & 0.052 $\pm$ 0.018 & 0.065 $\pm$ 0.022\\
    0.1143 & 0.079 $\pm$ 0.055 & 0.052 $\pm$ 0.011 & 0.065 $\pm$ 0.017\\
    \hline
    \end{tabular}
    \label{tab:mean_dev_L1err_norms_TC1} 
\end{table}

\begin{table}
    \caption{Mean and deviation of $L_\infty$ error norm: Test Case 1}
    \centering
    \begin{tabular}{cccc}
    \hline
    $t/T$ & $\lambda=0$ & $\lambda=8.5\times {10}^{-5}$ & $\lambda=1.7\times {10}^{-4}$\\\hline
    0.0127 & 0.165 $\pm$ 0.052 & 0.123 $\pm$ 0.005 & 0.125 $\pm$ 0.005\\
    0.0381 & 0.235 $\pm$ 0.152 & 0.149 $\pm$ 0.019 & 0.177 $\pm$ 0.028\\
    0.0635 & 0.287 $\pm$ 0.242 & 0.168 $\pm$ 0.074 & 0.242 $\pm$ 0.076\\
    0.0889 & 0.306 $\pm$ 0.238 & 0.195 $\pm$ 0.070 & 0.227 $\pm$ 0.071\\
    0.1143 & 0.235 $\pm$ 0.175 & 0.161 $\pm$ 0.043 & 0.192 $\pm$ 0.063\\
    \hline
    \end{tabular}
    \label{tab:mean_dev_L1inferr_norms_TC1} 
\end{table}

The mean error norms are somewhat different than the error norms presented in Figs.~\ref{fig:pred_errs_case1} (a) and (b), although the overall behavior is similar. For both the error norms, the plain LSTM has large deviations compared to the mean values. This indicates that it is a non-trivial task to obtain a plain LSTM network that can provide good predictions for both test cases. Both the physics-based LSTM are more reliable as they have much smaller deviations from their mean value. Also, for all the error norms for all available cases, the mean error norms of the network with $\lambda=8.5\times {10}^{-5}$ are significantly lower than their counterparts for the plain LSTM.

\subsection{Effects of physical regularizer}
Here we will further analyze the results presented earlier to explain the effects of the physical regularizer in significantly improving the predictions of the physics-based LSTM. We will specifically focus on Test Case 1 for brevity as the general behavior was found similar for both test cases. 

\begin{figure}[ht]
    \centering
    \begin{subfigure}{.49\textwidth}
        \centering
        \includegraphics[width=\linewidth]{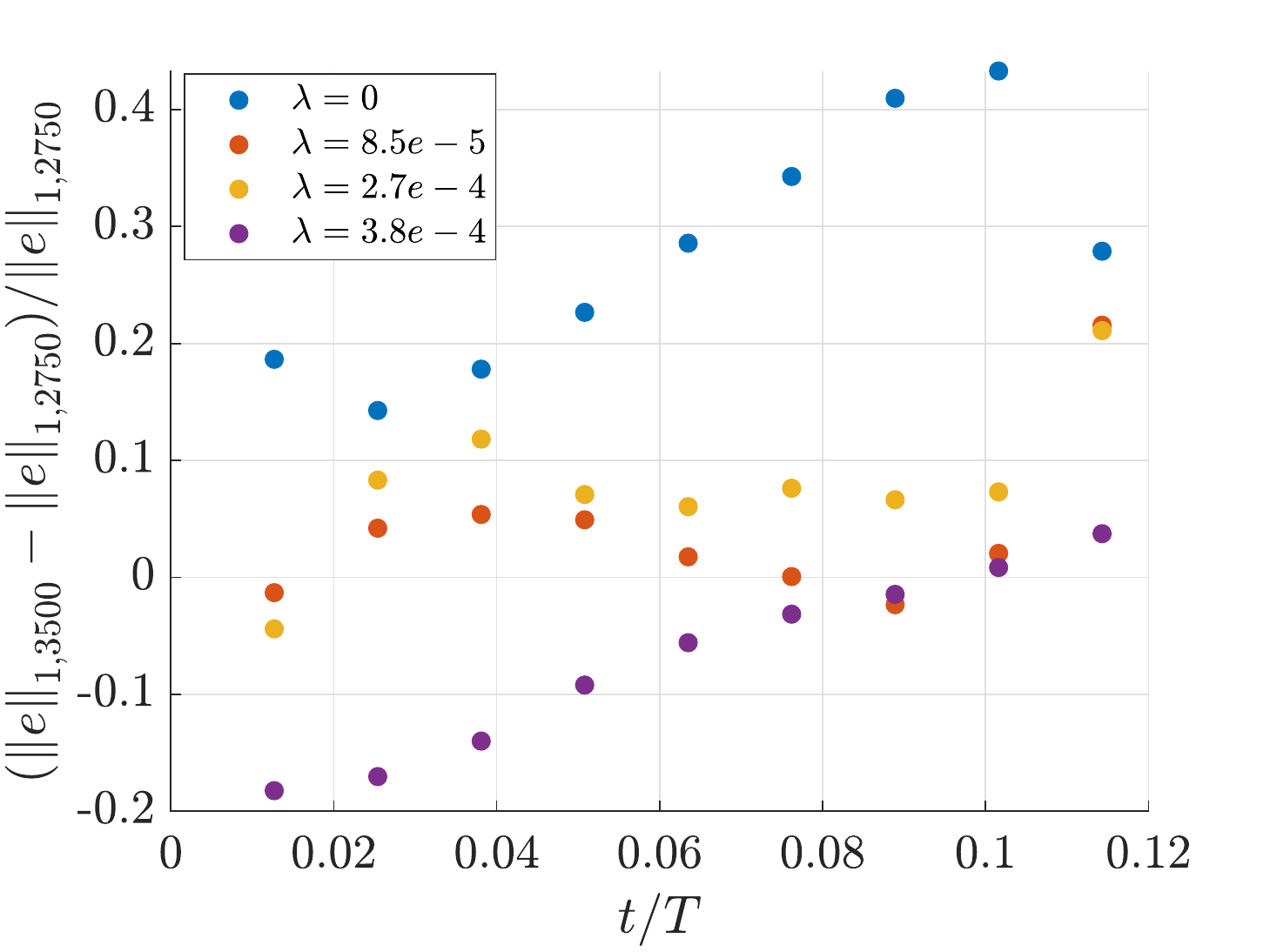}
        \caption{Relative $L_1$ error}
    \end{subfigure}
    \hfill
    \begin{subfigure}{.49\textwidth}
        \centering
        \includegraphics[width=\linewidth]{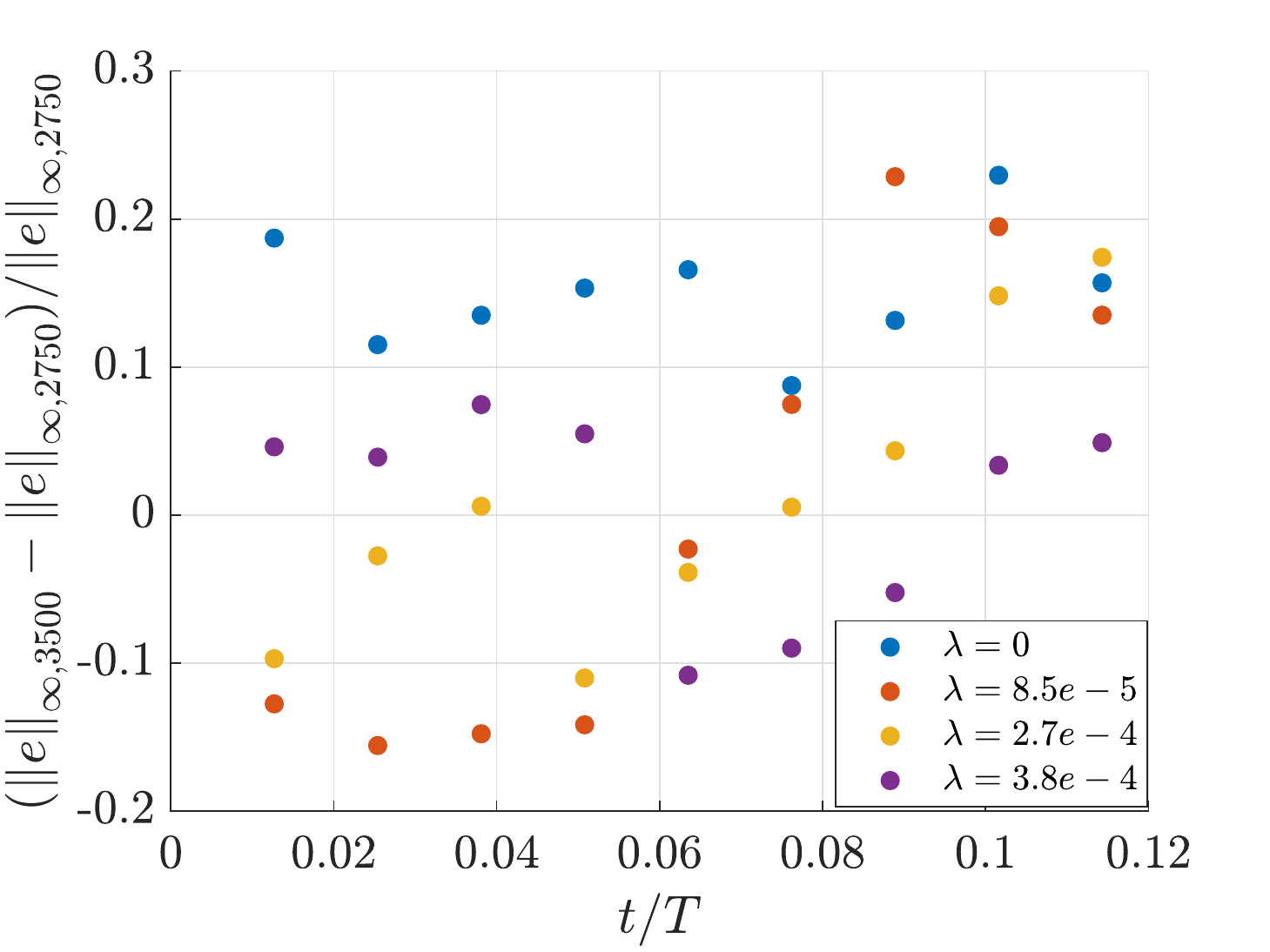}
        \caption{Relative $L_\infty$ error}
    \end{subfigure}
    \caption{Overfitting errors between 2750 epochs and 3500 epochs trained networks}
    \label{fig:overfit_red_reg}
\end{figure}

One of the main motivations behind using a regularizer is to prevent overfitting, where reduction in the training error by increasing the number of training epochs has an adverse effect on test/forecasting predictions. To study such overfitting effects, predictions for Test Case 1 were obtained for all LSTM networks trained for 3500 epochs and their corresponding early-stopped versions with 2750 training epochs. The relative normalized $L_1$ and $L_\infty$ error norms of the 3500 epoch-trained networks computed with respect to their 2750 epoch counterparts are presented in Fig.~\ref{fig:overfit_red_reg}. Fig.~\ref{fig:overfit_red_reg} (a) shows that the plain LSTM predictions experience an almost 15-20\% increase in $L_1$ error for $t=0.0127T$ to $t=0.0381T$ into the horizon and the relative $L_1$ error increases to almost 45\% for $t=0.1016T$. On the other hand, the physics-based LSTMs show a much smaller increase in the $L_1$ error with an increase in $\lambda$, and the relative $L_1$ error actually decreases for $\lambda=3.8\times {10}^{-4}$. The plain LSTM shows a smaller relative increase in the $L_\infty$ error norm (Fig.~\ref{fig:overfit_red_reg} (b)) compared to the $L_1$ error, but these increments are still larger overall than their physics-based counterparts. The relative  $L_\infty$ error norms of the physics-based LSTM do not change monotonically with $\lambda$ and vary with the elements of the sequence. Overall, $\lambda=2.7\times {10}^{-4}$ shows the least effect of overfitting.

\begin{figure}[ht]
    \centering
    \begin{subfigure}{.49\textwidth}
        \centering
        \includegraphics[width=\linewidth]{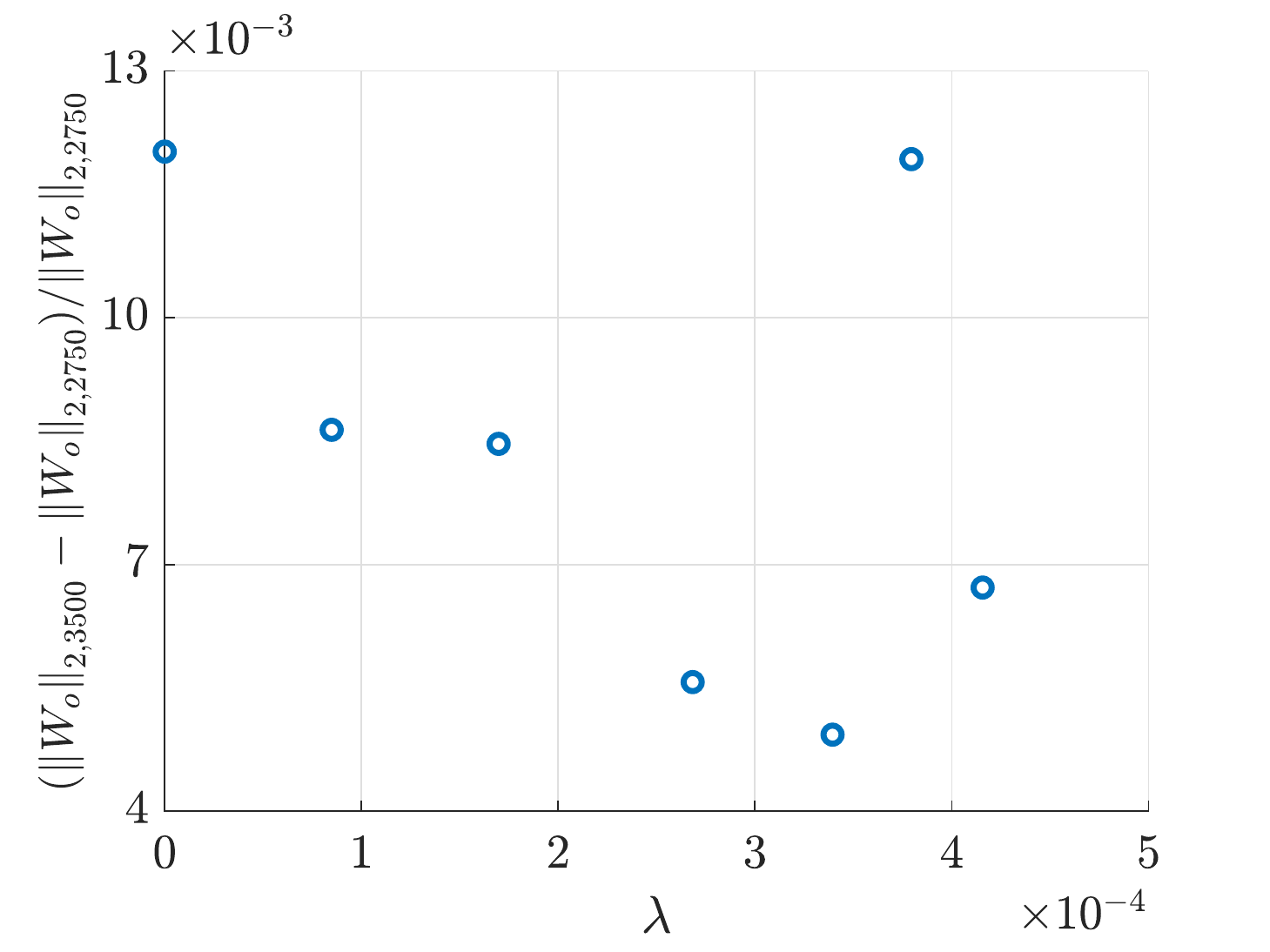}
        \caption{Relative change in ${\Vert W_o \Vert}_2$}
    \end{subfigure}
    \hfill
    \begin{subfigure}{.49\textwidth}
        \centering
        \includegraphics[width=\linewidth]{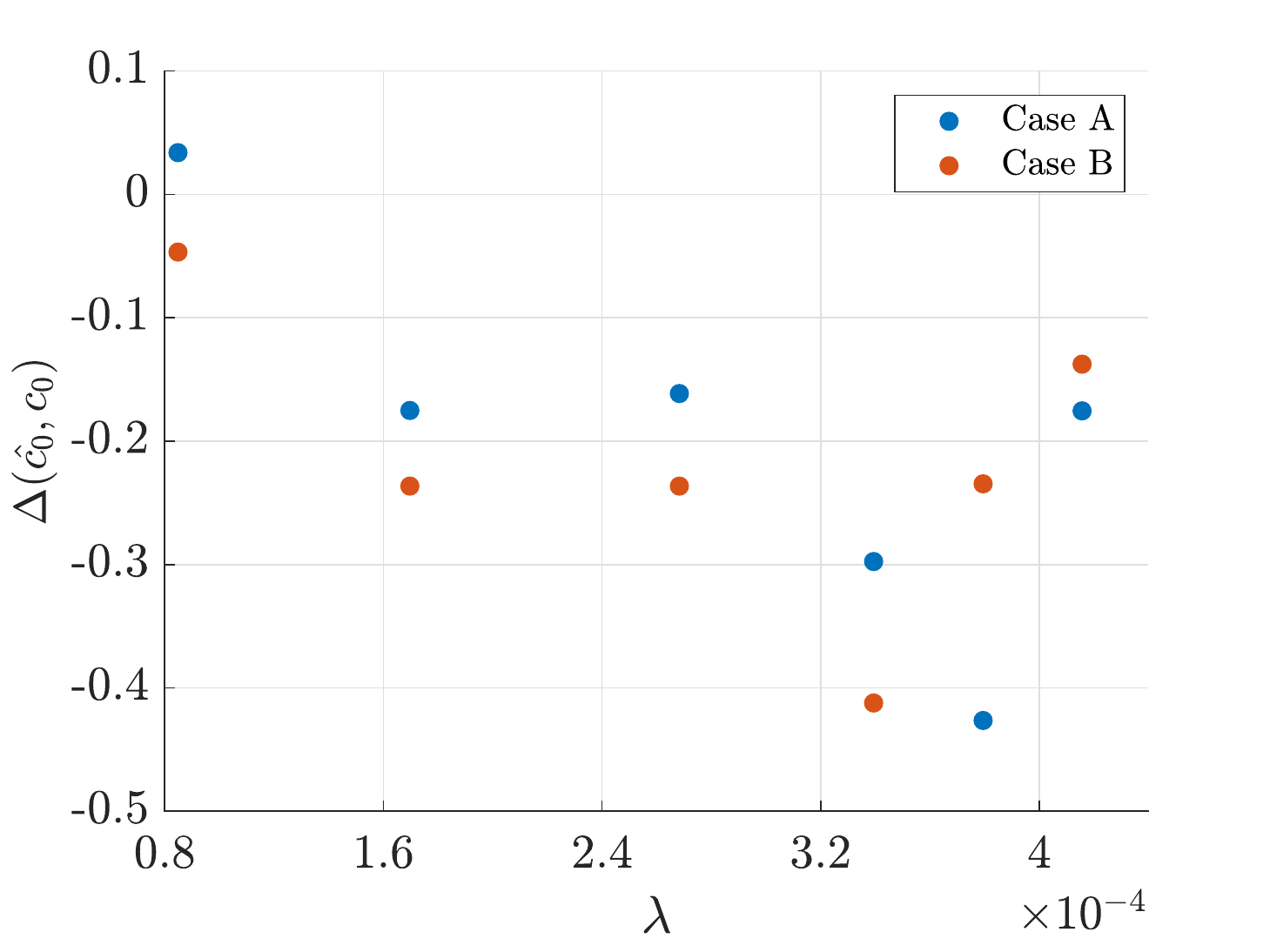}
        \caption{Relative wave speed $L_2$ errors}
    \end{subfigure}
    \caption{Effects of physical regularizer on wave speed and weight increment}
    \label{fig:effect_phys_reg}
\end{figure}

Regularizers usually reduce overfitting by reducing the weight increments with the increasing training epochs without significantly affecting the biases. This serves as a mean to reduce variance and provide a better variance-bias trade-off. To detect a similar behavior in the application of the present physical regularizer, we compute the relative change in the $L_2$ norm of the weights for LSTM networks trained for 3500 epochs compared to their counterparts trained for 2750 epochs. The relative change ${\Vert W \Vert}_2$ was normalized with the $L_2$ norm of the 2750 epoch training weights and the results for the LSTM output gate are shown in Fig.~\ref{fig:effect_phys_reg} (a). The results show that there is 1.1\% weight increment for the plain LSTM ($\lambda=0$) but for the physics-based LSTMs the weight increment goes down with an increase in the value of the hyperparameter $\lambda$ and the best results are obtained for $\lambda=2.7\times {10}^{-4}$, where the weight increment is only 5.8\%. The only anomaly to this behavior is the network with $\lambda=3.8\times {10}^{-4}$.

Here, we want to interpret the physical significance of our physical regularizer by relating the artificial bulk modulus of the hyperparameter to the normalized wave speed prediction errors for the various LSTM networks. For this, the normalized $L_2$ norm of the differences in wave speed from the training predictions of various physics-based LSTM networks, trained for 3500 epochs, are computed for cases A and B. For Case A, the normalized differences are computed relative to the $L_2$ errors of training wave speed of the early-stopped (trained for 2750 epochs) physics-based LSTM networks. For Case B, the normalized differences are computed relative to the $L_2$ errors of training wave speed training wave speed of plain LSTM networks ($\lambda=0$) trained for 3500 epochs. These can also be expressed as: 
\begin{gather}
   \Delta (\hat{c_0},c_0)= 
    \begin{cases}
        \Big[ \Vert \hat{c_0}-c_0 \Vert_2 - \Vert \hat{c_0}-c_0 \Vert_{2,2750}\Big] \slash \Vert \hat{c_0}-c_0 \Vert_{2,2750} \quad \text{for Case A}, \\
        \Big[ \Vert \hat{c_0}-c_0 \Vert_2 - \Vert \hat{c_0}-c_0 \Vert_{2,\lambda=0}\Big] \slash \Vert \hat{c_0}-c_0 \Vert_{2,\lambda=0} \quad \text{for Case B},
    \end{cases}
\end{gather}
where $\Delta (\hat{c_0},c_0)$ denotes the normalized difference in the $L_2$ errors of wave speed, and $\Vert \hat{c_0}-c_0 \Vert_2$ denote the $L_2$ errors in wave speed of the physics-based networks with 3500 training epochs.

The results shown in Fig.~\ref{fig:effect_phys_reg} (b) indicate that the wave speed $L_2$ errors of physics-based LSTMs decrease with continued training and the largest decrease of almost 42\% is obtained for $\lambda=3.8\times {10}^{-4}$. However, since this value of $\lambda$ also corresponds to the highest weight increment amongst the physics-based networks, the network weights may actually be slightly overfitted to reduce the wave speed $L_2$ errors. This can explain the dual effect of lowest wave speed $L_2$ errors and highest weight increment amongst the physics-based networks. However, even the physics-based networks with other values of $\lambda$ showed a reduction in the $L_2$ errors of wave speed on increasing the training epochs from 2750 to 3500. When the wave speed $L_2$ errors of the 3500 epoch-trained physics-based networks was compared to their plain LSTM counterpart (Case B), it also showed an overall reduction of the $L_2$ errors. A 41\% reduction was observed for $\lambda=3.4\times {10}^{-4}$. On increasing $\lambda$, the relative reduction of the wave speed $L_2$ errors of the physics-based LSTM does not decrease any further. The reason being that the main component of our loss function is the $L_2$ errors of the predicted pressure fluctuations, and the wave speed computations actually depend on the accuracy of the pressure predictions. Beyond the optimal range presented in Fig~\ref{fig:rel_pred_errs_case1} and discussed in the previous subsection, the accuracy of the predictions decreases, also leading to larger inaccuracies in the characteristic patterns of the predictions. Overall, the optimal values of the $\lambda$, the artificial bulk modulus, lies somewhere between $\lambda=1.6\times {10}^{-4}$ to $\lambda=3.2\times {10}^{-4}$. In this range, predictions obtained from the networks are not only expected to show smaller error norms but also provide the most consistent wave characteristic patterns.

\section{Conclusions and future work}
Here a kinematically consistent LSTM network was presented to learn inverse problems for one-dimensional wave propagation phenomena. Kinematical consistency is induced into the network via a physical regularizer, which minimizes the mean square errors of wave speed computed from target and predicted training data. The regularizer is scaled by a hyperparameter analogous to an artificial bulk modulus and is expected to provide more physically interpretable results. To the best of the authors' knowledge, such an application of kinematical consistency for inverse learning of mechanistic problems is novel in the physics-guided ML community, especially for wave propagation phenomena.

The results presented in this article show that for an optimal range of the hyperparameter $\lambda=8.5\times 10^{-5}$ to $\lambda=1.7\times 10^{-4}$, a 45\% reduction in the $L_1$ error and a 20\% reduction in the $L_\infty$ error of the plain LSTM predictions is possible via the physics-based LSTM. However, both the $L_1$ and $L_\infty$ error norms reduce by almost 50\% as we proceed further into the horizon. This enables the optimally tuned physics-based LSTM to forecast accurately almost two times further into the horizon compared to the plain LSTM. Furthermore, the optimally tuned physics-based LSTM also provides more robust predictions as indicated by their lower deviation of the $L_1$ and $L_\infty$ errors about the mean values compared to the plain LSTM networks.

The physics-based LSTMs' improved performance can be attributed to their efficacy in reducing overfitting. This is demonstrated by the smaller increase in the $L_1$ and $L_\infty$ errors of the test predictions of the physics-based LSTMs with an increase in the training duration, compared to the plain LSTM. Also, the physics-based LSTMs show a much smaller increment of the network weights with an increase in the training duration overall, compared to the plain LSTM. The physical regularization effect presented here can be related to the reduction in the $L_2$ error of the wave speed for the physics-based LSTMs compared to the plain LSTM. The smallest $L_2$ errors of wave speed prediction from the training target and predicted signals lie in the range of $\lambda=1.6\times 10^{-4}$ to $\lambda=3.2\times 10^{-4}$. This suggests that when the hyperparameter, resembling an artificial bulk modulus, lies in such a range, the magnitude of prediction errors can be reduced while maintaining the consistency of the wave characteristic patterns. The optimal range based purely on the smallest forecasting error norms is a subset of this range and thus further supports this deduction.

The present research, where we employ kinematical consistency as physical priors, serves as our proof of concept for integrating such physical information to more complex deep learning frameworks for learning two/three-dimensional wave propagation phenomena. The simplicity and generality of the proposed approach lie in its data-driven application of physical priors without resorting to the underlying PDEs anywhere. Future work would explore the enhancement in the scalability provided by such physical inductions when we try to learn more complex wave propagation phenomena, which better idealizes the underwater ocean environment. We will also comment further on the potential limitations of the proposed approach while employing them in presence of more complex wave interference phenomena.  

\bibliographystyle{ieeetr} 
\bibliography{neurips_2021_refs}

\begin{thebibliography}{10}

\bibitem{hsieh2009machine}
W.~W. Hsieh, {\em Machine learning methods in the environmental sciences:
  Neural networks and kernels}.
\newblock Cambridge university press, 2009.

\bibitem{bergen2019machine}
K.~J. Bergen, P.~A. Johnson, V.~Maarten, and G.~C. Beroza, ``Machine learning
  for data-driven discovery in solid earth geoscience,'' {\em Science},
  vol.~363, no.~6433, 2019.

\bibitem{reichstein2019deep}
M.~Reichstein, G.~Camps-Valls, B.~Stevens, M.~Jung, J.~Denzler, N.~Carvalhais,
  {\em et~al.}, ``Deep learning and process understanding for data-driven earth
  system science,'' {\em Nature}, vol.~566, no.~7743, pp.~195--204, 2019.

\bibitem{mansour2019deep}
T.~Mansour, ``Deep neural networks are lazy: on the inductive bias of deep
  learning,'' Master's thesis, Massachusetts Institute of Technology, 2019.

\bibitem{bronstein2021geometric}
M.~M. Bronstein, J.~Bruna, T.~Cohen, and P.~Veli{\v{c}}kovi{\'c}, ``Geometric
  deep learning: Grids, groups, graphs, geodesics, and gauges,'' {\em arXiv
  preprint arXiv:2104.13478}, 2021.

\bibitem{taddei2020registration}
T.~Taddei, ``A registration method for model order reduction: data compression
  and geometry reduction,'' {\em SIAM Journal on Scientific Computing},
  vol.~42, no.~2, pp.~A997--A1027, 2020.

\bibitem{mojgani2020physics}
R.~Mojgani and M.~Balajewicz, ``Physics-aware registration based auto-encoder
  for convection dominated pdes,'' {\em arXiv preprint arXiv:2006.15655}, 2020.

\bibitem{greif2019decay}
C.~Greif and K.~Urban, ``Decay of the kolmogorov n-width for wave problems,''
  {\em Applied Mathematics Letters}, vol.~96, pp.~216--222, 2019.

\bibitem{miyanawala2019decomposition}
T.~P. Miyanawala and R.~K. Jaiman, ``Decomposition of wake dynamics in
  fluid--structure interaction via low-dimensional models,'' {\em Journal of
  Fluid Mechanics}, vol.~867, pp.~723--764, 2019.

\bibitem{cybenko1989approximation}
G.~Cybenko, ``Approximation by superpositions of a sigmoidal function,'' {\em
  Mathematics of control, signals and systems}, vol.~2, no.~4, pp.~303--314,
  1989.

\bibitem{hornik1991approximation}
K.~Hornik, ``Approximation capabilities of multilayer feedforward networks,''
  {\em Neural networks}, vol.~4, no.~2, pp.~251--257, 1991.

\bibitem{leshno1993multilayer}
M.~Leshno, V.~Y. Lin, A.~Pinkus, and S.~Schocken, ``Multilayer feedforward
  networks with a nonpolynomial activation function can approximate any
  function,'' {\em Neural networks}, vol.~6, no.~6, pp.~861--867, 1993.

\bibitem{pinkus1999approximation}
A.~Pinkus, ``Approximation theory of the mlp model,'' {\em Acta Numerica 1999:
  Volume 8}, vol.~8, pp.~143--195, 1999.

\bibitem{gonzalez2018deep}
F.~J. Gonzalez and M.~Balajewicz, ``Deep convolutional recurrent autoencoders
  for learning low-dimensional feature dynamics of fluid systems,'' 2018.

\bibitem{sorteberg2019approximating}
W.~E. Sorteberg, S.~Garasto, C.~C. Cantwell, and A.~A. Bharath, ``Approximating
  the solution of surface wave propagation using deep neural networks,'' in
  {\em INNS Big Data and Deep Learning conference}, pp.~246--256, Springer,
  2019.

\bibitem{cheng2020data}
M.~Cheng, F.~Fang, C.~C. Pain, and I.~Navon, ``Data-driven modelling of
  nonlinear spatio-temporal fluid flows using a deep convolutional generative
  adversarial network,'' {\em Computer Methods in Applied Mechanics and
  Engineering}, vol.~365, p.~113000, 2020.

\bibitem{lee2020model}
K.~Lee and K.~T. Carlberg, ``Model reduction of dynamical systems on nonlinear
  manifolds using deep convolutional autoencoders,'' {\em Journal of
  Computational Physics}, vol.~404, p.~108973, 2020.

\bibitem{parish2020time}
E.~J. Parish and K.~T. Carlberg, ``Time-series machine-learning error models
  for approximate solutions to parameterized dynamical systems,'' {\em Computer
  Methods in Applied Mechanics and Engineering}, vol.~365, p.~112990, 2020.

\bibitem{bukka2021assessment}
S.~R. Bukka, R.~Gupta, A.~R. Magee, and R.~K. Jaiman, ``Assessment of unsteady
  flow predictions using hybrid deep learning based reduced-order models,''
  {\em Physics of Fluids}, vol.~33, no.~1, p.~013601, 2021.

\bibitem{willard2020integrating}
J.~Willard, X.~Jia, S.~Xu, M.~Steinbach, and V.~Kumar, ``Integrating
  physics-based modeling with machine learning: A survey,'' {\em arXiv preprint
  arXiv:2003.04919}, 2020.

\bibitem{jia2019physics}
X.~Jia, J.~Willard, A.~Karpatne, J.~Read, J.~Zwart, M.~Steinbach, and V.~Kumar,
  ``Physics guided rnns for modeling dynamical systems: A case study in
  simulating lake temperature profiles,'' in {\em Proceedings of the 2019 SIAM
  International Conference on Data Mining}, pp.~558--566, SIAM, 2019.

\bibitem{raissi2019physics}
M.~Raissi, P.~Perdikaris, and G.~E. Karniadakis, ``Physics-informed neural
  networks: A deep learning framework for solving forward and inverse problems
  involving nonlinear partial differential equations,'' {\em Journal of
  Computational Physics}, vol.~378, pp.~686--707, 2019.

\bibitem{yang2020physics}
L.~Yang, D.~Zhang, and G.~E. Karniadakis, ``Physics-informed generative
  adversarial networks for stochastic differential equations,'' {\em SIAM
  Journal on Scientific Computing}, vol.~42, no.~1, pp.~A292--A317, 2020.

\bibitem{daw2020physics}
A.~Daw, R.~Q. Thomas, C.~C. Carey, J.~S. Read, A.~P. Appling, and A.~Karpatne,
  ``Physics-guided architecture (pga) of neural networks for quantifying
  uncertainty in lake temperature modeling,'' in {\em Proceedings of the 2020
  siam international conference on data mining}, pp.~532--540, SIAM, 2020.

\bibitem{sun2020theory}
J.~Sun, Z.~Niu, K.~A. Innanen, J.~Li, and D.~O. Trad, ``A theory-guided
  deep-learning formulation and optimization of seismic waveform inversion,''
  {\em Geophysics}, vol.~85, no.~2, pp.~R87--R99, 2020.

\bibitem{ling2016reynolds}
J.~Ling, A.~Kurzawski, and J.~Templeton, ``Reynolds averaged turbulence
  modelling using deep neural networks with embedded invariance,'' {\em Journal
  of Fluid Mechanics}, vol.~807, pp.~155--166, 2016.

\bibitem{wang2020incorporating}
R.~Wang, R.~Walters, and R.~Yu, ``Incorporating symmetry into deep dynamics
  models for improved generalization,'' {\em arXiv preprint arXiv:2002.03061},
  2020.

\bibitem{udrescu2020ai}
S.-M. Udrescu and M.~Tegmark, ``Ai feynman: A physics-inspired method for
  symbolic regression,'' {\em Science Advances}, vol.~6, no.~16, p.~eaay2631,
  2020.

\bibitem{ruthotto2019deep}
L.~Ruthotto and E.~Haber, ``Deep neural networks motivated by partial
  differential equations,'' {\em Journal of Mathematical Imaging and Vision},
  pp.~1--13, 2019.

\bibitem{greydanus2019hamiltonian}
S.~Greydanus, M.~Dzamba, and J.~Yosinski, ``Hamiltonian neural networks,'' {\em
  arXiv preprint arXiv:1906.01563}, 2019.

\bibitem{toth2019hamiltonian}
P.~Toth, D.~J. Rezende, A.~Jaegle, S.~Racani{\`e}re, A.~Botev, and I.~Higgins,
  ``Hamiltonian generative networks,'' {\em arXiv preprint arXiv:1909.13789},
  2019.

\bibitem{wang2020towards}
R.~Wang, K.~Kashinath, M.~Mustafa, A.~Albert, and R.~Yu, ``Towards
  physics-informed deep learning for turbulent flow prediction,'' in {\em
  Proceedings of the 26th ACM SIGKDD International Conference on Knowledge
  Discovery \& Data Mining}, pp.~1457--1466, 2020.

\bibitem{wehmeyer2018time}
C.~Wehmeyer and F.~No{\'e}, ``Time-lagged autoencoders: Deep learning of slow
  collective variables for molecular kinetics,'' {\em The Journal of chemical
  physics}, vol.~148, no.~24, p.~241703, 2018.

\bibitem{otto2019linearly}
S.~E. Otto and C.~W. Rowley, ``Linearly recurrent autoencoder networks for
  learning dynamics,'' {\em SIAM Journal on Applied Dynamical Systems},
  vol.~18, no.~1, pp.~558--593, 2019.

\bibitem{chang2019antisymmetricrnn}
B.~Chang, M.~Chen, E.~Haber, and E.~H. Chi, ``Antisymmetricrnn: A dynamical
  system view on recurrent neural networks,'' {\em arXiv preprint
  arXiv:1902.09689}, 2019.

\bibitem{ba2019blending}
Y.~Ba, G.~Zhao, and A.~Kadambi, ``Blending diverse physical priors with neural
  networks,'' {\em arXiv preprint arXiv:1910.00201}, 2019.

\bibitem{duarte2021soundscape}
C.~M. Duarte, L.~Chapuis, S.~P. Collin, D.~P. Costa, R.~P. Devassy, V.~M.
  Eguiluz, C.~Erbe, T.~A. Gordon, B.~S. Halpern, H.~R. Harding, {\em et~al.},
  ``The soundscape of the anthropocene ocean,'' {\em Science}, vol.~371,
  no.~6529, 2021.

\bibitem{tensorflow2015-whitepaper}
M.~Abadi, A.~Agarwal, P.~Barham, E.~Brevdo, Z.~Chen, C.~Citro, G.~S. Corrado,
  A.~Davis, J.~Dean, M.~Devin, S.~Ghemawat, I.~Goodfellow, A.~Harp, G.~Irving,
  M.~Isard, Y.~Jia, R.~Jozefowicz, L.~Kaiser, M.~Kudlur, J.~Levenberg,
  D.~Man\'{e}, R.~Monga, S.~Moore, D.~Murray, C.~Olah, M.~Schuster, J.~Shlens,
  B.~Steiner, I.~Sutskever, K.~Talwar, P.~Tucker, V.~Vanhoucke, V.~Vasudevan,
  F.~Vi\'{e}gas, O.~Vinyals, P.~Warden, M.~Wattenberg, M.~Wicke, Y.~Yu, and
  X.~Zheng, ``{TensorFlow}: Large-scale machine learning on heterogeneous
  systems,'' 2015.
\newblock Software available from tensorflow.org.

\end{thebibliography}




\end{document}